\RequirePackage{fix-cm}
\documentclass[twocolumn]{svjour3}          % twocolumn
\smartqed  % flush right qed marks, e.g. at end of proof
\usepackage{graphicx}
\usepackage{multirow}
\usepackage{cite}
\usepackage{color}
\usepackage{amssymb}
\usepackage{amsmath}
\usepackage{mathtools}
\usepackage{subcaption}
\usepackage{xcolor}
\usepackage{url}
\usepackage{booktabs}
\usepackage{enumitem}
\usepackage[T1]{fontenc}
\usepackage[rightcaption]{sidecap}
\usepackage[normalem]{ulem}
\usepackage{soul}
\usepackage{makecell}

\sidecaptionvpos{figure}{c}
\usepackage[pagebackref=true,breaklinks=true,letterpaper=true,colorlinks,bookmarks=false]{hyperref}
\usepackage{tabularx}

\begin{document}
\emergencystretch 3em
\title{HWNet v2: An Efficient Word Image Representation for Handwritten Documents.}
%\subtitle{Do you have a subtitle?\\ If so, write it here}

%\titlerunning{Short form of title}        % if too long for running head

\author{Praveen Krishnan         \and
        C.V. Jawahar %etc.
}

%\authorrunning{Short form of author list} % if too long for running head

\institute{Praveen Krishnan \at
              CVIT, IIIT Hyderabad, India \\
            %   Tel.: +XXX-XX-XXXXXX\\
            %   Fax: ++XXX-XX-XXXXXX\\
              \email{praveen.krishnan@research.iiit.ac.in}           %  \\
%             \emph{Present address:} of F. Author  %  if needed
           \and
           C.V. Jawahar \at
           CVIT, IIIT Hyderabad, India \\
           \email{jawahar@iiit.ac.in} 
}

\date{Received: date / Accepted: date}
% The correct dates will be entered by the editor

\maketitle

\begin{abstract}
We present a framework for learning an efficient holistic representation for handwritten word images. The proposed method uses a deep convolutional neural network with traditional classification loss. The major strengths of our work lie in: (i) the efficient usage of synthetic data to pre-train a deep network, (ii) an adapted version of the ResNet-34 architecture with the region of interest pooling (referred to as HWNet v2) which learns discriminative features for  variable sized word images, and (iii) a realistic augmentation of training data with multiple scales and distortions which mimics the natural process of handwriting. We further investigate the process of transfer learning to reduce the domain gap between synthetic and real domain, and also analyze the invariances learned at different layers of the network using visualization techniques proposed in the literature.

Our representation leads to a state-of-the-art word spotting performance on standard handwritten datasets and historical manuscripts in different languages with minimal representation size. On the challenging \textsc{iam} dataset, our method is first to report an mAP of around $0.90$ for word spotting with a representation size of just $32$ dimensions. Furthermore, we also present results on printed document datasets in English and Indic scripts which validates the generic nature of the proposed framework for learning word image representation.

% \PACS{PACS code1 \and PACS code2 \and more}
% \subclass{MSC code1 \and MSC code2 \and more}
\end{abstract}

\section{Introduction}
\label{sec:intro}
%Information extraction from the documents has revolutionized the way we gather knowledge on distinct topics. These documents include books, manuscripts and numerous other data such as letters, invoices, catalogs which are either machine printed or handwritten. Over the years, many of these documents have been digitized with an aim to provide better access, preservation of historical manuscripts and to enable language technologies which could be integrated into modern information processing frameworks. The key to achieving this goal lies in learning an efficient representation of document images and its constituent parts, from pixel space to feature space, which is rich in preserving both lexical and semantic information. In this work, we focus on handwritten document images and, address the challenges in learning such a representation for efficient content level access. More specifically, we set our granularity at the level of word images which could be taken as the fundamental semantic unit of a document image.

%digitization of printed and written documennts have opened lot of possibilities such as widespread access , preservation, language pipeline
%technology for accessing it at content level is not mature enough. One of the key reasons for this is that feature representations are not go
%Good feature representation is able to  preserve both lexical and semantic information.
%In this work we focus on representing word images in a feature space which preservs lexical information.

Digitization of documents has opened numerous possibilities in providing widespread access to information and creating data for building language processing pipelines such as machine translation and search engines. These digitized documents are either machine printed or handwritten and include books, manuscripts, letters, invoices, catalogs etc. Content-level access to this extensive digital corpus is possible only by learning an efficient representation of document images which is rich in preserving both lexical and semantic information. In this work, we set our granularity at the level of words, and focus on representing word images in a feature space which preserves its lexical information.

Content level access can be achieved either through ``recognition'' or ``retrieval''. Given the advancements in the field of information retrieval and language processing, one can build robust and innovative solutions using the text produced from an ideal recognizer. However, the document images that we are interested in this work are handwritten, historical manuscripts and degraded printed books where traditional printed Optical Character Recognizer (\textsc{ocr}) based methods would result in noisy text and could lead to inferior results. This leads us to the complementary method, where the idea is to formulate the problem from a ``retrieval' perspective in either recognition-based~\cite{ToselliHMM2016, FischerHMM2013} or recognition-free~\cite{ManmathaHR96,AlmazanPAMI14, RusinolATL11} manner. In this setting, given a query word image, one has to rank all the word images from the candidate set in the order of its similarity. This idea was popularized as `word spotting' in ~\cite{ManmathaHR96}. Here both the query and the candidate word images are represented using a holistic representation which captures the lexical information of a word in an appropriate feature space. The design of these holistic representations is one of the major challenges to be solved, which decides the effectiveness of recognition-free methods. In this work, we propose a new holistic descriptor for word images which can seamlessly be used for developing applications involving handwritten and printed documents. We assume the segmentation of document images into words is given to us either in the form of ground truth or made available using external word proposal method which could be noisy. The proposed representation in this work is a word-level descriptor.

Feature engineering has been a key investigation for any pattern recognition problem. In the domain of document images, the problem of defining an optimal feature which describes a word~\cite{ManmathaHR96,AlmazanPAMI14,AldavertRTL15,Rusinol15}, character/patch~\cite{rodriguez2008local,terasawa2009slit,Roy2015} has been an interesting quest in the community over the last two decades. With the improved representations over time, there has been a significant impact on the larger goals of the community such as recognition and retrieval of documents, script recognition, layout analysis, etc. In the domain of word images, initial features proposed were based on pixel level statistics~\cite{MartiB01,RathM03} which worked only on limited settings of the writers and font variations. Later, with the popularization of local gradient level features in computer vision such as scale-invariant feature transform (\textsc{sift})~\cite{Lowe04} and histogram of gradients (\textsc{hog})~\cite{dalal2005histograms}, are adapted to document images due to their generic properties. These features are invariant to scale, translation, and common degradation. The bag of visual words (\textsc{bow})~\cite{csurka2004visual,sivic2003video}, were built using these local features along with advanced encoding schemes such as Fisher~\cite{perronnin2007fisher}, locality constrained linear coding (\textsc{llc})~\cite{wang2010locality}, and sparse codes~\cite{yang2009linear}. These features along with the learned models such as~\cite{AlmazanPAMI14}, obtained state-of-the-art word spotting and recognition for historical manuscripts and multi-writer handwritten documents.

\begin{figure}[t]
\centering
\includegraphics[width=7.5cm]{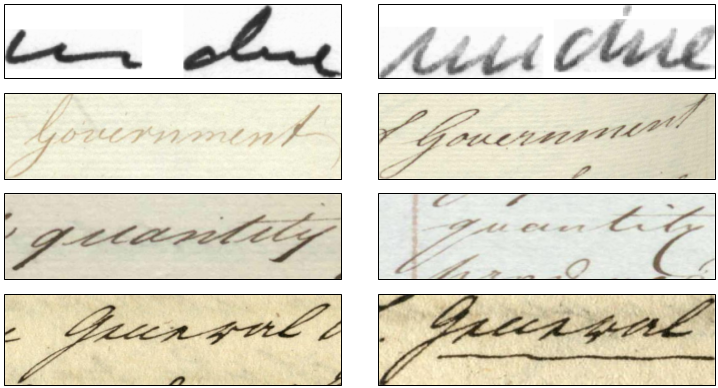}
\caption{Sample top-1 nearest neighbors in the learned representation space. Here we show examples from historical datasets which contains degradation and irregular segmentation of words. One can also notice invariance of representation in terms of handwriting variations as shown in the top row.}
\label{fig:illus}
\end{figure}

More recently, there has been a paradigm shift from `feature engineering' to `feature learning' due to the resurgence of neural networks. It is mostly credited to the revival of convolutional neural networks (\textsc{cnn})~\cite{KrizhevskySH12}, availability of large scale of annotated data~\cite{imagenet_cvpr09}, and increased computing power using graphical processing units (\textsc{gpu}). The features are learned on the fly during training and gets adapted to the task of interest. Networks trained on large data sets also learn generic feature representation that can be used for related tasks~\cite{RazavianASC14}, and in many cases, have reported state-of-the-art results as compared to the handcrafted features. When the data is limited, fine-tuning a pre-trained network has also been demonstrated to be very effective. In the domain of document images these features have shown better performance for word spotting~\cite{krishnan2016matching,SudholtF16,wilkinson2016semantic,Sudholt2018}, recognition~\cite{Poznanski_2016_CVPR}, document classification~\cite{harley2015evaluation}, layout analysis~\cite{chen2015page}, etc. In this work, we propose a deep \textsc{cnn} architecture named as HWNet v2, for the task of learning an efficient word level representation for handwritten documents which can handle multiple writers and, is robust to common forms of degradation and noise. We also show the generic nature of the proposed representation and architecture which allows it to be used as an off-the-shelf feature for printed documents and in building state-of-the-art word spotting systems for various languages. In order to derive a compact representation for an efficient storage and retrieval, we evaluate the performance of compressed feature codes, where we push the compression to an order of 16-to-32 dimensions with minimal drop in performance. Fig.~\ref{fig:illus} shows sample word images which are considered as nearest neighbors in the proposed representation space. The shown images are quite challenging in terms of handwriting variations, distortion created due to scanning, and irregular segmentation which are common in historical manuscripts. These images are taken from the test sets of the datasets used in this work, which are explained more in detail in Section~\ref{sec:exp}.

\subsection{Contributions} 
The baseline \textsc{cnn} architecture HWNet considered in this work was first proposed in~\cite{krishnan2016matching}, which first demonstrated the use of such an architecture in learning an efficient word level representation. This work is dedicated entirely to enrich the representation space and learning better in-variances that are common to the handwritten data. The major contributions of this work are: (i) bringing architectural improvements by making it more deep with the help of residual connections, referred as HWNet v2, (ii) use of multi-scale training and region of interest pooling (\textsc{roi}) to support variable sized word images, (iii) use of extensive augmentation scheme with elastic distortion which is better suited for handwritten images. In addition to this, we perform an in-depth analysis of transfer learning at various layers and also visualize the features at different layers of the network to get better insights on the learned features and its invariances. We also significantly compress our representations for efficient storage and faster retrieval. We further extend our experiments on newer datasets from historical documents and degraded printed books from Indic scripts and present state-of-the-art results on standard datasets.

\begin{figure}[t]
\includegraphics[width=8.2cm]{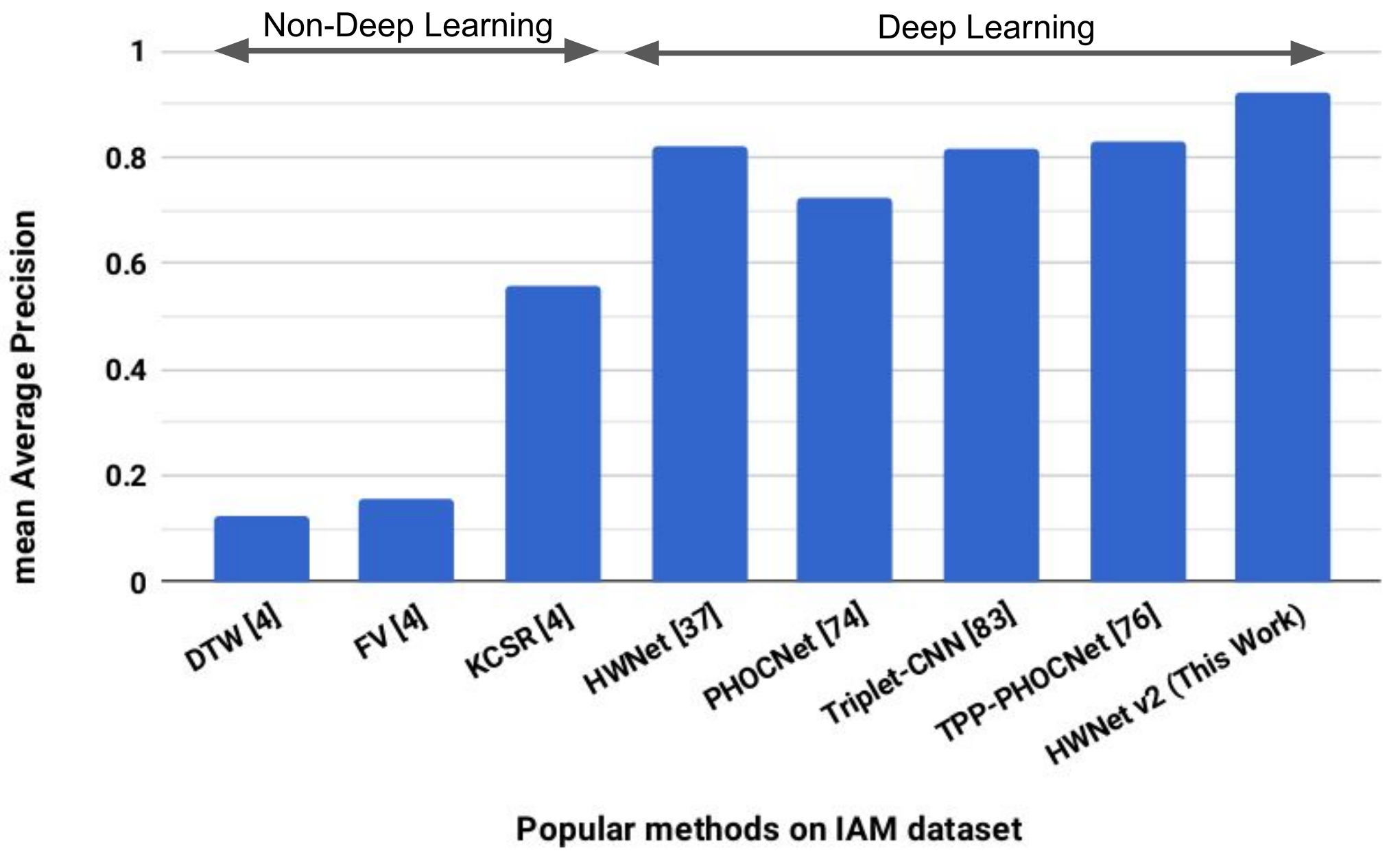}
\caption{Evolution of word spotting methods from the perspective of different word image representation schemes. The evaluation is conducted on the IAM~\cite{Marti02} dataset using mean Average Precision (mAP). More details on evaluation scheme and dataset is provided in Section~\ref{subsec:datEval}.}
\label{fig:method-perf}
\end{figure}

\section{Related Works}
\label{sec:relWorks}
There are numerous successful attempts done in the past which focus on different aspects of the problem such as the modality of data (printed, handwritten and scene text), nature of representation either fixed or variable length, and type of embedding scheme. We broadly split our discussion on related works into three major parts: (i) the classical methods built using variable length representation schemes, (ii) fixed length representation which are achieved using bag of words framework, and (iii) learned representation using, different classifier models built on top of handcrafted features and using the deep learning networks. Fig~\ref{fig:method-perf} presents an evolution of some of the key methods proposed in the space of word spotting, which uses some of the recent representations proposed for handwritten word images. As one can notice, with the introduction of deep learning, there has been a significant boost in the performance. Our proposed method, which is shown in the end further boosts the performance in this space. In Table~\ref{tab:relWorks}, we present a quick overview of the related works that we discuss in this section based on its representation type. One can also refer to the detailed survey presented in~\cite{GiotiPR17} which reviews major representation schemes proposed for the task of word spotting.

%Note that both HWNet~\cite{krishnan2016matching} and PHOCNet~\cite{SudholtF16} methods were published around the same time.
% \renewcommand{\arraystretch}{1.2}
\begin{table}[b]
\centering
\caption{Overview of major methods proposed in the literature, summarizing different types of word image representation schemes.}
\label{tab:relWorks}
\begin{tabular}{|l|l|}\hline
Method & Representation \\\hline
Profile~\cite{RathM03} & \multirow{4}{*}{Variable Length} \\
Profile+Moments~\cite{MartiB01} & \\
Profile+DFT~\cite{AKumar07} & \\
Slit HoG~\cite{terasawa2009slit} & \\\hline
Local Gradient Histogram~\cite{rodriguez2008local} & \multirow{2}{*}{HMM} \\
SIFT~\cite{RothackerICDAR13} & \\\hline
SIFT~\cite{aldavert2013integrating, RusinolATL11, AldavertRTL15, ShekharJ12, Yalniz12} & BoWs \\\hline
SIFT~\cite{AlmazanPR14} & Fisher \\\hline
SIFT~\cite{AlmazanPAMI14} & PHOC Attributes \\\hline
Deep Learning~\cite{krishnan2016matching} & Neural Codes \\\hline
Deep Learning~\cite{praveenICFHR16,SudholtF16, sudholt2017evaluating,wilkinson2016semantic} & PHOC Attributes \\\hline
Deep Learning~\cite{gomezlsde2017} & Levenshtein Embedding \\\hline
\end{tabular}
\end{table}

\subsection{Classical Representation}
\label{subsec:classicalRep}
Learning holistic representation for word images was popularized as word spotting which was originally proposed in~\cite{Rohlicek89} for speech processing. Within the document community, initial attempts in this space mostly focused on variable length representations of word images by considering it as a temporal sequence. Most of these methods used profile features~\cite{ManmathaHR96,MartiB01, RathM03} which are computed at each column of the word image and are summarized using various pixel level statistics. Dynamic Time Warping (\textsc{dtw}) based algorithms were found to be useful for matching variable length representations and is quite popular in speech~\cite{Sakoe78, Myers80} and other sequence matching problems. In~\cite{RathM03}, Rath et al. used  profile features namely vertical profile, upper \& lower word profile, and background to ink transitions. In~\cite{Bala06, MesheshaJ08}, profile features were combined with the shape based structural features for a partial matching scheme using \textsc{dtw}. Although these features are fast to compute, it is susceptible to noise and common degradation present in documents.

With the popularity of the local gradient features such as \textsc{sift}~\cite{Lowe04}, \textsc{hog}~\cite{dalal2005histograms} which describes a patch using histograms of edge orientations computed from a gradient image, the features are less susceptible to stray pixels and variations in brightness and contrast. Methods such as~\cite{terasawa2009slit,rodriguez2008local} adapted local gradient features for word spotting where~\cite{terasawa2009slit} used a continuous \textsc{dtw} algorithm for partial word matching from the line images and~\cite{rodriguez2008local} used Hidden Markov Model (\textsc{hmm}) based classification method. Most of the features discussed above are not robust to different fonts, writing styles and required careful image pre-processing techniques such as binarization, slant and skew correction which remain hard for handwritten and historical documents. Moreover, the methods such as \textsc{dtw} and \textsc{hmm} based scheme of matching variable length representations do not scale to large datasets due to the higher time complexity. Hence, the later methods appreciated more on fixed length representations built on top of highly engineered features proposed in computer vision.

\subsection{Bag of Word Representation}
\label{subsec:bowRep}
The popularity of bag of words (\textsc{bow})~\cite{csurka2004visual,sivic2003video} framework using local gradient features such as \textsc{sift} and \textsc{hog}, led to its proliferation to document images~\cite{RusinolATL11,ShekharJ12,Yalniz12,aldavert2013integrating,Rusinol15,AldavertRTL15}. Rusinol et al.~\cite{RusinolATL11,Rusinol15}, presented a patch based framework using \textsc{bow} histograms computed from the underlying \textsc{sift} descriptors. The histogram based representation was further projected onto a topic space using latent semantic indexing (\textsc{lsi})~\cite{deerwester1990indexing}, where the latent topic space is assumed to preserve the lexical content of word images. In~\cite{ShekharJ12,Yalniz12}, \textsc{bow} based representation was adapted for word image retrieval for machine printed documents using standard keypoint detectors such as Harris~\cite{harris1988combined}, \textsc{fast}~\cite{rosten2006machine} corner detectors, while the local features at the keypoints were computed using \textsc{sift}. Due to the fixed length and sparse nature of the representation, the matching was done using cosine distance and an inverted index was used for faster retrieval. 
%To improve the precision of the search, Yalniz et al.~\cite{Yalniz12} proposed Longest Common Subsequence (\textsc{lcs}) based spatial verification scheme between the top-k retrieved results and the original query representation. 
Aldavert~\cite{AldavertRTL15} et al., presents a detailed survey for \textsc{bow} based representation for handwritten word spotting with its analysis on the effect of codebook size, choice of encoding, and type of normalization. In a similar line of work using local features and codebook, Almaz{\'{a}}n et al.~\cite{AlmazanPR14} uses an exemplar-\textsc{svm}~\cite{malisiewicz2011ensemble} for representing a query and performs the initial scoring of candidate words. Given the initial matches, the list is re-ranked using a Fisher vector based representation~\cite{PerronninR09} which is a generalization of \textsc{bow} using higher order statistics. In general, the unsupervised nature of learning of \textsc{bow} based methods make them directly applicable to historical databases where the annotation is hard and costly. However, the limitation of local features (\textsc{sift,hog}) to capture the larger part level information from word images restricted these methods to work only for limited writer datasets where the variations are less.

%In general, \textsc{bow} based methods are scalable in nature and due to the unsupervised nature of learning, the methods are directly applicable on historical databases where annotation is hard and costly.
\subsection{Learned Representations}
\label{subsec:learnedRep}
Learning representation in supervised settings leads to better discriminative features at the cost of annotations. Here, we present popular methods from literature which uses label embedding techniques together with traditional machine learning models and modern deep learning architectures.

\subsubsection{Word Attributes}
\label{subsubsec:wordAttr}
Almaz{\'{a}}n et al.~\cite{AlmazanPAMI14} proposed a label embedding approach where both word images and the corresponding labels are embedded into a common vectorial subspace which allows comparing both modalities seamlessly. The method uses a new textual representation referred to as pyramidal histogram of characters (\textsc{phoc}), which concatenates the histogram of characters at multiple spatial regions in a pyramidal fashion. Here, each feature denotes the presence or absence of a particular character at a particular spatial region and is called a character level attribute. Similar to the textual representation, \textsc{phoc} representation for word images could be derived from the scores of attribute level binary classifiers, which are learned from training word images which posses that particular attribute. In~\cite{AlmazanPAMI14}, Almaz{\'{a}}n et al. uses Fisher features of word images for learning word image attributes while \textsc{phoc} for text labels are extracted by the spatial position of each character. 
%Given {\sc phoc} representation for both image and text,~\cite{AlmazanPAMI14} uses common subspace regression technique for label embedding which essentially learns a common subspace where image and text can be matched uses nearest neighbors.A similar label embedding technique was also simultaneously proposed for scene text recognition using \textsc{phoc} like features in~\cite{rodriguez2015label} and uses structured SVM formulation for joint embedding. 
%\textcolor{red}{Although the word attribute framework is generic, the underlying features used are handcrafted which does not truly capture the spirit of holistic features by taking the entire context of word image into account at multiple resolutions.} 
Although the word attribute framework is generic, the underlying handcrafted features limit the robustness of the learned holistic features. In recent methods, this is addressed using deep neural networks to learn features in an end to end hierarchical manner which results in better generalization. 

% In the experiment section, we also show the improvement of such an attribute based framework when trained using the deep features proposed in this work.

\subsubsection{Deep Learning}
\label{subsubsec:deepLearning}
With the advancements in deep learning, there is a paradigm shift in feature engineering where features are now learned during the training process which customizes itself to the domain of training data. Among different types of neural networks, deep convolutional neural networks (\textsc{cnn})~\cite{KrizhevskySH12,SimonyanZ14a,SzegedyLJSRAEVR15} have revolutionized the way features are learned for specific tasks. In the domain of word images, Jaderberg et al.~\cite{JaderbergSVZ14,JaderbergECCV14,jaderberg2014IJCV}, proposed three different architecture models (char, nGram and dictionary words) for scene text recognition. Taking inspirations for word attributes, for handwritten images, Poznanski et al.~\cite{Poznanski_2016_CVPR} adapted VGGNet~\cite{SimonyanZ14a} for recognizing \textsc{phoc} attributes by having multiple parallel fully connected layers, each one predicting \textsc{phoc} attributes at a particular level. In similar spirits, different architectures~\cite{SudholtF16, praveenICFHR16, wilkinson2016semantic, Sudholt2018} were proposed using \textsc{cnn} networks which embed features into different textual embedding spaces defined by {\sc phoc}. In~\cite{SudholtF16}, Sudholt et al. proposes an architecture to directly embed image features to {\sc phoc} attributes by having sigmoid activation in the final layer and thereby avoiding multiple fully connected layers as presented in~\cite{Poznanski_2016_CVPR}. It is referred to as PHOCNet, which uses the final layer activation to derive a holistic representation for word spotting. In the later set of works~\cite{Sudholt2018,sudholt2017evaluating} from the same group, PHOCNet was adapted with temporal pooling layer (TPP-PHOCNet) and evaluated under different loss functions and optimization algorithms which further improved the word spotting performance. In~\cite{praveenICFHR16, praveenDAS18}, the features computed from HWNet~\cite{krishnan2016matching} are embedded into word attribute space by training attribute based {\sc svm} classifiers and projecting both image and textual attributes to a common subspace. In~\cite{wilkinson2016semantic}, the authors propose a two stage architecture where a triplet \textsc{cnn} network is trained to reduce the distance between the anchor word image and a similar labeled (positive) word image, while simultaneously increasing the distance between the anchor and negative labeled word image. In the second stage, the learned image representation is embedded into a word embedding space (\textsc{phoc}, \textsc{dctow}, ngram etc) using a fully connected neural network. In the above methods where the target embedding is an attribute space, one can query the representation space either using query-by-string or query-by-example setting.
%Most of the above works uses the output activation from the penultimate layer of the \textsc{cnn} network as the word representation to perform spotting and retrieval.

In general, most of these methods learn representation restricted to a fixed attribute space whereas, in this work, our aim is to learn a generic representation for word images derived by formulating a proxy task of word image classification. We derive our representation from the penultimate layer before the softmax which takes the advantage of the hierarchical composition of concepts in a deep network where higher layers tend to capture parts and attribute level information. In terms of architecture advancements, we use  a much deeper network using residual connections and present variable sized images without distorting the aspect ratio of word images. The learned representation can be used for word attribute learning~\cite{praveenICFHR16,praveenDAS18}, spotting, and retrieval tasks for both handwritten and printed documents irrespective of the vocabulary used while training.

\subsection{Segmentation-Free Approaches}
\label{subsec:segFreeRelWorks}
In addition to the different representation schemes, one can also classify the methods in terms of segmentation-based and segmentation-free word spotting approach. Most of the methods presented so far belong to the setting where the segmentation of words is available in the form of ground truth. In segmentation-free setting, the input is a page image and the underlying method first proposes potential word hypothesis before computing its representation. In literature, one can place the segmentation-free approaches into three broad categories. The first category of methods~\cite{RothackerICDAR13, RusinolATL11, AlmazanPR14, ghosh2015sliding} use a sliding window technique where the regions are proposed along a regular grid. This typically results in a dense extraction of bounding boxes and are computationally expensive to process. The second category of methods utilize connected components~\cite{kovalchuk2014simple, ghosh2018text} along with mathematical morphological operations to extract characters/words from the page image. Most of these methods work in a bottom-up fashion and utilize certain rules to extract the final bounding box of words. The number of proposals obtained using this approach is far less than sliding windows, however, they are sensitive to page quality and degradation as seen in historical documents. In~\cite{rothacker2017word}, authors propose a hybrid approach where the document image is first subjected to dense text detection using sliding windows and later the word hypothesises are computed using the set of extremal regions. The third category of methods~\cite{wilkinson2017neural,wilkinsonArxiv18,axler2018toward} in the segmentation-free setting is inspired by the recent success of region proposal based object detection techniques such as Faster R-CNN~\cite{ren2015faster}. The Ctrl-F-Net~\cite{wilkinson2017neural} model proposes an end to end trainable detection and embedding network. It utilizes a localization layer to predict potential word proposals along with its wordness score. The initial predictions are further filtered and presented to the embedding network. The method also utilizes a complementary external region proposal method called as Dilated Text Proposals ({\sc dtp}) to improve the overall recall of the system. The authors extended their method in~\cite{wilkinsonArxiv18} by simplifying the architecture (Ctrl-F-Mini) by only utilizing the external proposals computed using {\sc DTP}. This performs faster and better in certain situations than the original architecture.

%In Section~\ref{subsubsec:segFreeResults} we demonstrate the robustness of the proposed representation of our work using the word proposals obtained from the Ctrl-F-Mini architecture.

\begin{figure*}[t]
\centering
\begin{subfigure}[b]{0.28\textwidth}
        \includegraphics[width=\textwidth]{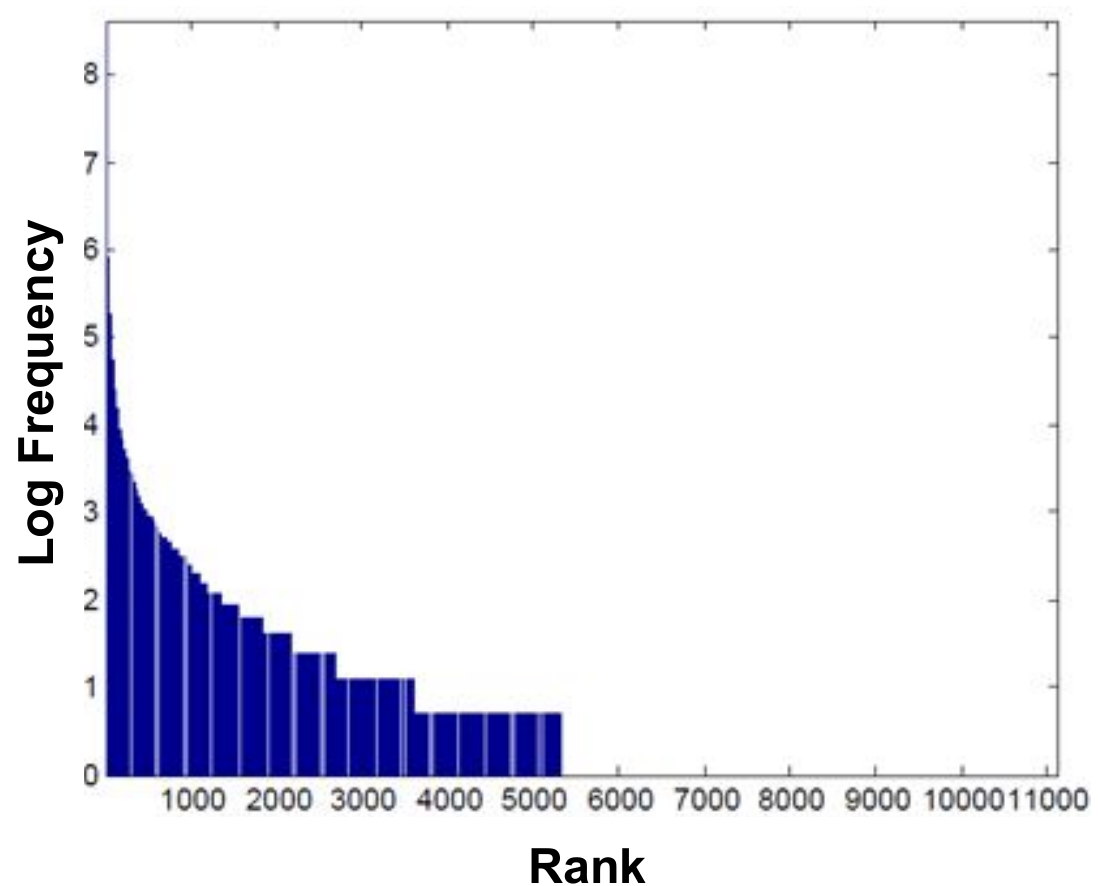}
        \caption{}
        \label{fig:iam-zipf}
    \end{subfigure}
\begin{subfigure}[b]{0.7\textwidth}
        \includegraphics[width=\textwidth]{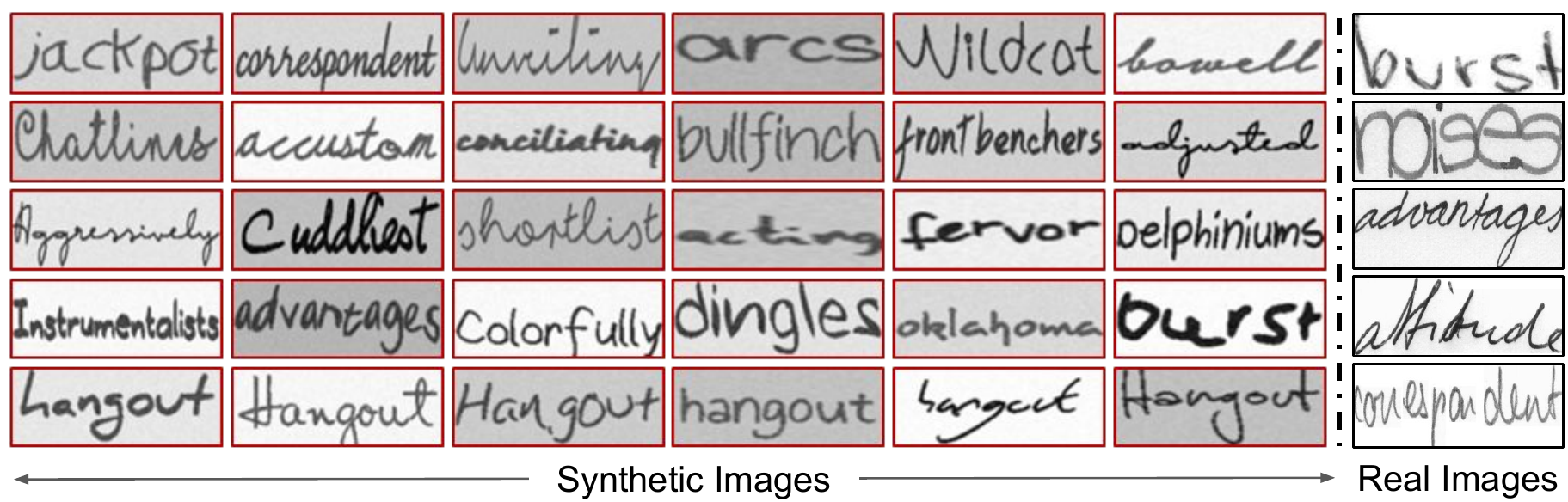}
        \caption{}
        \label{fig:iiit-hws-images}
    \end{subfigure}
\caption{(a) Distribution of words in \textsc{iam} dataset, (b) Sample word images from the {\sc iiit-hws} dataset created as part of this work, to address the lack of training data for learning deep \textsc{cnn} networks.}
\label{fig:wordImages}
\end{figure*}

\section{Handwritten Synthetic Dataset}
\label{sec:hwsynth}
Quality data~\cite{Everingham10,imagenet_cvpr09} has always played a pivotal role in the advancement of pattern recognition problems. Some of the key properties for any dataset are: (i) a good sample distribution which mimics the real world unseen examples, (ii) quality of annotation, and (iii) scale. With the success of deep learning based methods~\cite{KrizhevskySH12,SimonyanZ14a,SzegedyLJSRAEVR15,JaderbergSVZ14}, there has been a surge in newer supervised learning architectures which are ever more data hungry. These architectures have millions of parameters to learn, thereby need a large amount of training data to avoid over-fitting and to generalize well. In general, data creation is a time consuming and expensive process which requires huge human efforts. More recently, an alternative form of data generation process with minimal supervision is getting popular~\cite{JaderbergSVZ14,RozantsevLF15,ros2016synthia}, which uses synthetic mechanisms to render and annotate images in an appropriate form. The simple idea of generating data synthetically allows overcoming the challenges in obtaining the data. In this work, we address the need for large scale annotated datasets for handwritten images by generating synthetic words with natural variations. Fig.~\ref{fig:wordImages} (b) shows sample handwritten word images generated using the proposed framework which looks quite natural and comparable with its counterparts from the real world which are shown in the last column of the figure.

Some of the popular datasets in handwritten domain are \textsc{iam}  handwriting dataset~\cite{Marti02}, George Washington~\cite{Fischer12, ManmathaHR96}, Bentham manuscripts~\cite{causer2012building}, Parzival database~\cite{Fischer12} etc. Except for \textsc{iam}, the remaining datasets are part of the historical collections which were created by one or very few writers. \textsc{iam} is a relatively modern dataset, which consists of unconstrained text written in forms by around 657 writers. The vocabulary of \textsc{iam} is limited to nearly 11K words whereas any normal dictionary in the English language would contain more than 100K words. Fig.~\ref{fig:wordImages} (a) shows the distribution of entire words in \textsc{iam} vocabulary which follows the typical Zipf law. As one can notice that, out of 11K words, nearly 10.5K word classes contain fewer than 20 samples or instances. Also, the majority of remaining words are stop words which are shorter in length and are less informative. The actual samples in training data are much smaller than this, which limits building efficient deep learning networks such as~\cite{KrizhevskySH12}.

\subsection{Handwritten Font Rendering}
\label{subsec:fontRender}
We use publicly available handwritten fonts for our task. The vocabulary of words is chosen from a dictionary. For each word in the vocabulary, we randomly sample a font and render\footnote{We use ImageMagick for rendering the word images. URL: \url{http://www.imagemagick.org/script/index.php}} its corresponding image. During this process, we vary the following parameters: (i) kerning level (inter character space), (ii) stroke width, from a defined distribution. In order to make the pixel distribution of both foreground ($F_g$) and background ($B_g$) pixels more natural, we sample the corresponding pixels, for both regions from a Gaussian distribution where the parameters such as mean and standard deviation are learned from the $F_g$ and $B_g$ region of \textsc{iam} dataset. Finally, Gaussian filtering is done to smooth the rendered image.

\subsection{IIIT-HWS Dataset}
\label{subsec:iiithws}
To address the lack of data for training handwritten word images for English, we build a synthetic handwritten dataset of 1 million word images. We call this dataset as {\sc iiit-hws}. Some of the sample images from this dataset are shown in Fig.~\ref{fig:wordImages} (b). Note that these images are very similar to natural handwriting. The {\sc iiit-hws} dataset is formed out of 750 publicly available handwritten fonts. We use the popular Hunspell dictionary and pick a unique set of 90K words for this purpose. For each word, we randomly sample 100 fonts and render its corresponding image. Moreover, we prefer to learn a case-insensitive model for each word category, hence we perform three types of rendering, namely, all letters capitalized, all letters lower, and only the first letter in caps.

\begin{figure}[t]
\centering
\includegraphics[height=4.5cm,width=8.3cm]{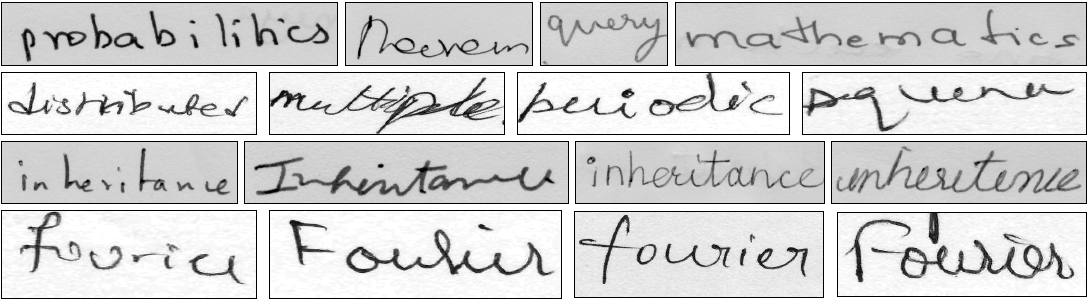}
\caption{Top two rows show the variations in handwritten images, the bottom two rows demonstrate the challenges of intra class variability in images across writers.}
\label{fig:hwChallenges}
\end{figure}

\section{HWNet}
\label{sec:hwnet}
In the quest for learning better holistic features for word images, we leveraged recent \textsc{cnn} architectures to learn discriminative representations. Fig.~\ref{fig:hwChallenges} demonstrates the challenges across writers. The top two rows show the variations across images in which some are even hard for humans to read without enough context of nearby characters. The bottom two rows show different instances of the same word written by the different writers, e.g., ``inheritance'' and ``Fourier'', where one can clearly notice the variability in shape for each character in the word image. The learned representation needs to be invariant to (i) both inter and intra class variability across the writers, (ii) presence of skew, (iii) quality of ink, and (iv) quality and resolution of the scanned image. One can also notice that there can be instances where few characters are completely distorted or degraded due to the cursive nature of word formation. However, the knowledge of vocabulary and its overall appearance, humans can still make out the word. One of the key differentiation of word images with respect to natural scene images is that a word image is inherently a variable length representation and making it fixed size would distort the individual characters non-uniformly.

We propose a deep \textsc{cnn} architecture named as HWNet, first presented in~\cite{krishnan2016matching} for learning representation for handwritten word images, and further enhance the network capacity to address issues specific to handwriting. The improved network architecture is one of the major contributions of this work which is named as HWNet v2. We formulate the problem as word classification on a given vocabulary, however, given a trained network, we are interested to derive holistic features $f\in\mathbb{R}^d$ which obeys lexical similarity to all possible words in that language (irrespective of the trained vocabulary).

\begin{figure}[t]
\centering
\begin{subfigure}[b]{0.5\textwidth}
        \includegraphics[width=\textwidth]{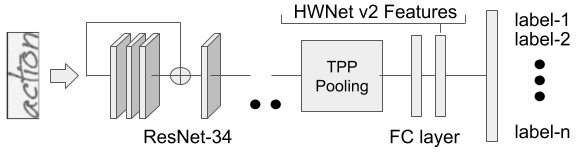}
        \caption{}
        \label{fig:hwnet-arch}
    \end{subfigure}
\begin{subfigure}[b]{0.5\textwidth}
        \includegraphics[width=\textwidth]{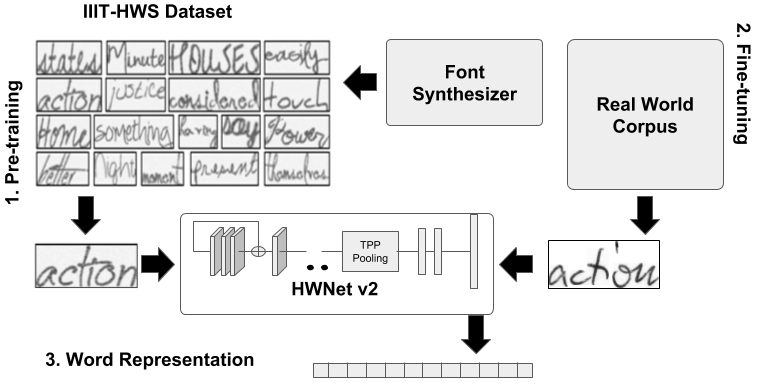}
        \caption{}
        \label{fig:hwnet-finetune}
    \end{subfigure}
\caption{(a) HWNet v2 architecture which comprises of a deep CNN architecture using ResNet blocks along with a TPP pooling and fully connected layers, (b) Flowchart showing the transfer learning process where we first pre-train the network on synthetic data and later fine-tune it on real corpus. The features are extracted from the penultimate layer of the network.}
\label{fig:hwnetArch}
\end{figure}

\subsection{HWNet Baseline Architecture}
\label{subsec:arch}
In the baseline model, we use a fixed sized gray scale word image of dimension $48\times128$. As noted earlier, this would result in distortion of aspect ratio, however, we show in experiments, the learned features are robust, and quite better than previous variable size representations. The underlying architecture of our \textsc{cnn} model is inspired from~\cite{KrizhevskySH12}. We use a~\textsc{cnn} with five convolutional layers with 64, 128, 256, 512, and 512 square filters with dimensions: 5, 5, 3, 3 and 3 respectively. The next two layers are fully connected ones with 2048 neurons each. The last layer uses a fully connected (FC) layer with dimension equal to the number of classes (vocabulary of training and validation set), and is further connected to the softmax layer to compute the class specific probabilities. Rectified linear units (ReLU) are used as the non-linear activation units after each weight layer except the last one, and $2 \times 2$ max pooling is applied after first, second, and fourth convolutional layers. We use a stride of one and padding is done to preserve the spatial dimensionality. We empirically observed that using batch normalization~\cite{IoffeS15} after each convolutional and fully connected layer, resulted in lower generalization error as compared to dropouts. We use cross entropy loss function to predict the word class labels, and the weights are updated using the mini batch gradient descent algorithm with momentum.

\setlength\tabcolsep{1pt}
\begin{table}[t]
\centering
\caption{Summary of the HWNet v2 network configuration. The width, height, and number of channels of each convolution layer are shown in square brackets, with the number of layers that are stacked together. We present two variations in the network (as shown in the sixth column), using a single level {\sc roi} pooling or using temporal pyramid pooling {\sc (tpp)} with three levels.}
\label{tab:arch}
\begin{tabular}{|c|c|c|c|c|c|c|}
\hline
Conv1 & Block1 & Block2 & Block3 & Block4 & ROI/TPP & FC \\ \hline
3x3 & \makecell{{[}3x3,64{]} \\ x 3} & \makecell{{[}3x3,128{]} \\ x 4} & \makecell{{[}3x3,256{]}\\ x 6} & \makecell{{[}3x3,512{]} \\x 3} & \makecell{{ROI\{6x12\}} \\ TPP\{1,2,3\}} &\makecell{{[2048]} \\x 2} \\ \hline
\end{tabular}
\end{table}
\setlength\tabcolsep{2pt}

\section{HWNet v2}
\label{sec:hwnet++}
In our original HWNet architecture, we limited the number of convolutional layers to five, which was equivalent to layers proposed in AlexNet. More recently, newer architectures such as VGGNet~\cite{SimonyanZ14a}, GoogLeNet~\cite{SzegedyLJSRAEVR15}, and ResNets~\cite{he2016deep} have shown deeper \textsc{cnn} networks for better performance and the resulting features to be more discriminative. Some of the key architectural changes brought in these networks, which lead to efficient training are: (i) use of lower dimensional filters ($3 \times 3$) thereby having less parameters from larger sized filters, which also acts as a forced regularizer, (ii) use of ($1 \times 1$) filters which acts as dimensionality reduction unit to keep the no. of parameters in control, (iii) use of inception layer~\cite{SzegedyLJSRAEVR15} which introduces multi-scale processing by having multiple parallel layers operating at different scales, and (iv) use of residual blocks~\cite{he2016deep} to learn residual function $\mathcal{F}(x):=\mathcal{H}(x)-x$. Here $\mathcal{H}(x)$ is the desired underlying mapping. A residual layer is typically implemented using a shortcut connection without any parameters. In our improved HWNet architecture, named as HWNet v2, we use the ResNet34~\cite{he2016deep} network with four blocks where each block contains multiple ResNet modules. Instead of using global average pooling (as proposed along with ResNet architecture~\cite{he2016deep}), we found fully connected layers in the end for learning better features from the penultimate layer. Table~\ref{tab:arch} shows the summary of HWNet v2 network configuration. Here each ResNet module consists of two convolutional layers and a shortcut connection to enable residual learning. There is no max pooling in the network and the spatial resolution is down sampled using a stride of 2 at the first convolutional layer of block 2, 3 and 4. As stated earlier, we also use batch normalization after each convolutional and fully connected layer except the last one. 

\begin{figure}[b]
\centering
\includegraphics[width=8.5cm]{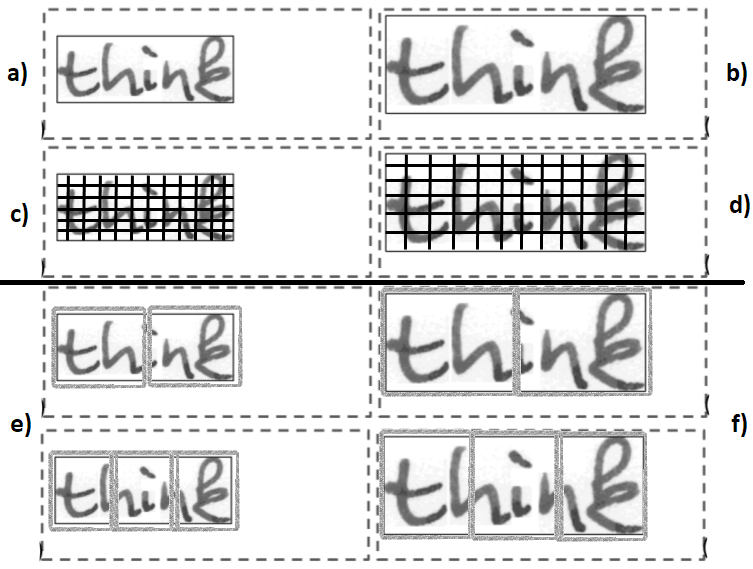}
\caption{(a-b) Multi-scale input, (c-d) region of interest pooling and, (e-f) temporal pyramid pooling shown at levels 2 and 3.}
\label{fig:multiscale}
\end{figure}

\subsection{Multi-Scale Training and ROI/TPP Pooling} 
\label{subsec:multiScale}
One of the major limitations in the previous architecture was the requirement to use fixed dimensional inputs so that it remains compatible with fully connected layers of the network. However as mentioned earlier, this leads to distortion of aspect ratio which manipulates the appearance of characters present in the word images arbitrarily. Another aspect which we tend to ignore so far due to the restriction of fixed size is, the ability to train word images in multiple scales so that the network remains invariant to multiple character scales. It has been observed that different writers typically write at different scales which leads to the presence of a variable sized sequence of characters. To overcome these issues, we use a fixed size padded image ($128\times 384$) to accommodate variable sized input word image. As shown in Fig.~\ref{fig:multiscale}(a-b), we render different scale input image to learn scale invariant representation. Given the output feature maps from the last convolutional layer, we only keep the activations coming from the input region belonging to word image using \textsc{roi} pooling. \textsc{roi} pooling~\cite{Girshick2015}, layer gives a differentiable pooling (max/average) mechanism from variable sized input feature maps into fixed sized output maps, by constructing a grid with variable sized cells. Fig.~\ref{fig:multiscale}(c-d), shows the \textsc{roi} pooling where the number of grids in both images remains same while the size of each grid cell varies as per the scale of the image. In another variant of the HWNet v2 network, we use temporal pyramid pooling {\sc (tpp)} similar to~\cite{sudholt2017evaluating}. Both {\sc roi} and {\sc tpp} takes variable length representation and produces a fixed length output depending on the number of grids. In our usage of {\sc tpp} along HWNet v2, we  follow a similar paradigm of creating pyramid levels only vertically~\cite{ShekharJ12, sudholt2017evaluating} with levels set at 1, 2 and 3. This would essentially capture the temporal properties present in the word image. Fig.~\ref{fig:multiscale}(e-f), shows the temporal pyramid pooling at level 2 and 3. We also set max pooling as the preferred pooling operation within each grid for both \textsc{roi} and \textsc{tpp} variants. In general, both \textsc{roi, tpp} pooling methods don't require a fixed sized padded image. We made such a decision from an implementation point of view so that we can train in batches of images with the fixed size which wouldn't be possible otherwise.

\begin{figure}[t]
\centering
\includegraphics[height=3.5cm]{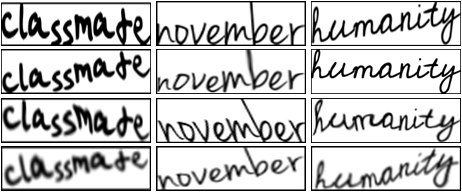}
\caption{Data augmentation techniques: affine and elastic distortion.}
\label{fig:dataAug}
\end{figure}

\subsection{Data Augmentation and Elastic Distortion} 
\label{subsec:dataAug}
While training a \textsc{cnn} network, data augmentation~\cite{KrizhevskySH12} is a common practice to introduce artificial variations in data to make network robust to intra-class variations and prevent over-fitting. Popular data augmentation techniques are random crops, horizontal reflection, random flipping of pixels, and affine transformations such as scaling and translation. In this work, while training HWNet v2, we perform two major augmentation schemes which are: (i) affine transformation and (ii) elastic distortion. In affine transformation, we generalize to translation, scaling, rotation, and shearing. Here rotation and shearing are restricted to certain angles which mimic the skew and cursiveness present in natural handwriting. Elastic distortion~\cite{SimardSP03} has been used in the past successfully for recognizing handwritten digits. It mimics variations created from the oscillation of hand and inertia exerted on the writing medium. The basic idea is to generate a random displacement field which dictates the computation of new location to each pixel through interpolation. The displacement field is smoothed using a Gaussian filter of standard deviation $\sigma$ and scaled using constant factor $\alpha$. Both $\sigma, \alpha$ are set empirically by visualizing the quality of distorted images. Fig.~\ref{fig:dataAug} shows different possible variations created for each word image.

\begin{figure*}[t]
\centering
\begin{subfigure}[b]{0.18\textwidth}
        \includegraphics[width=\textwidth]{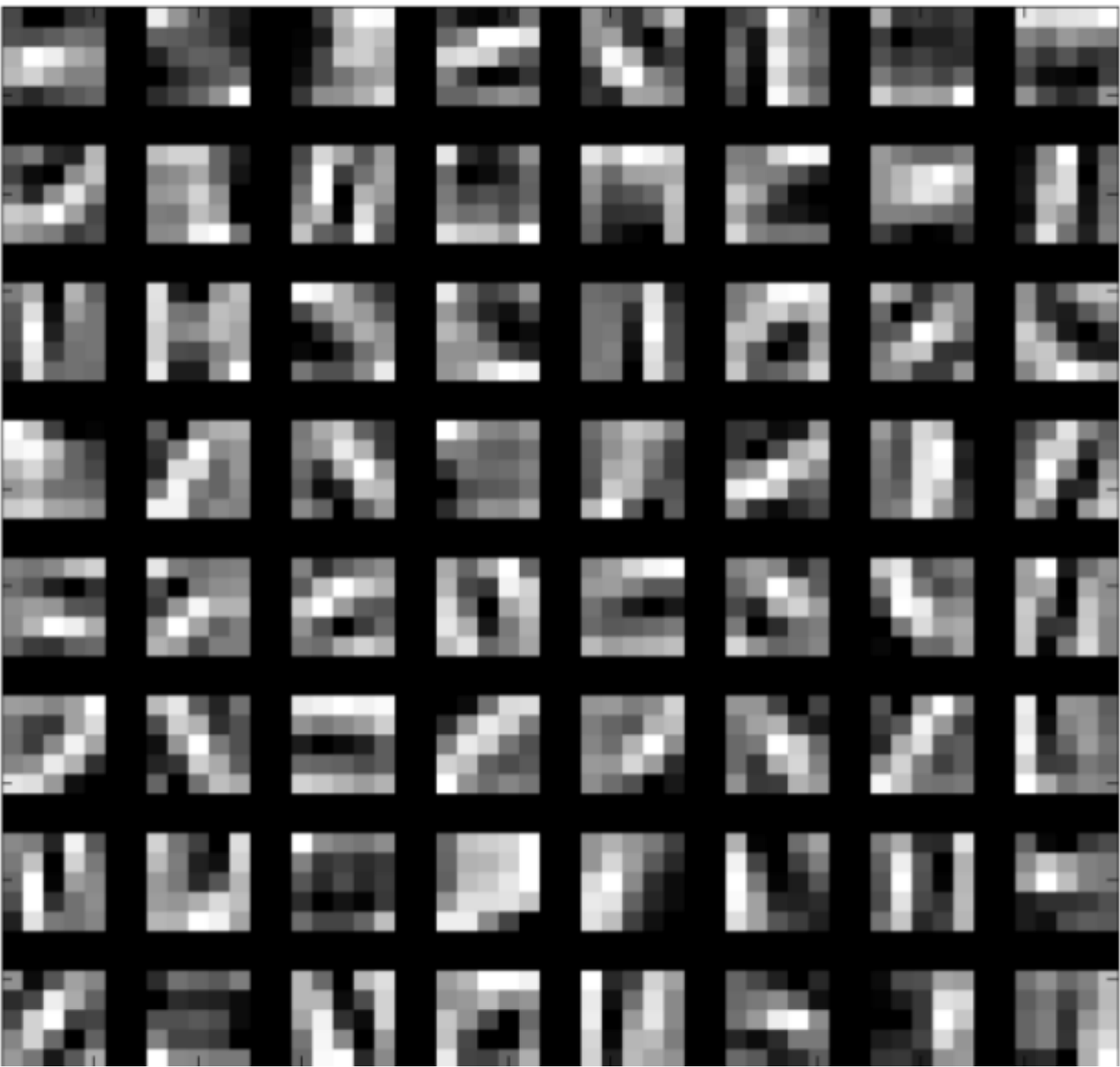}
        \caption{}
        \label{fig:visLayer1}
    \end{subfigure}
\begin{subfigure}[b]{0.52\textwidth}
        \includegraphics[width=\textwidth]{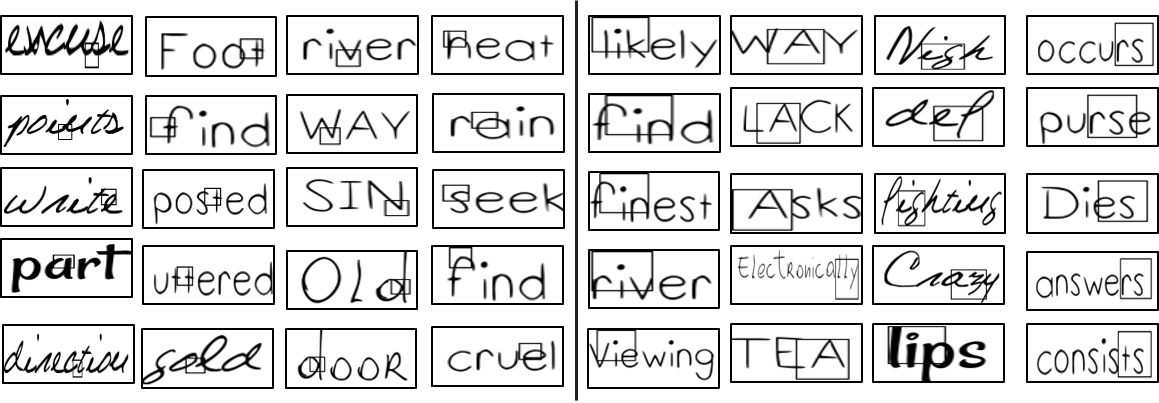}
        \caption{}
        \label{fig:visPatches}
    \end{subfigure}
\begin{subfigure}[b]{0.26\textwidth}
        \includegraphics[width=\textwidth]{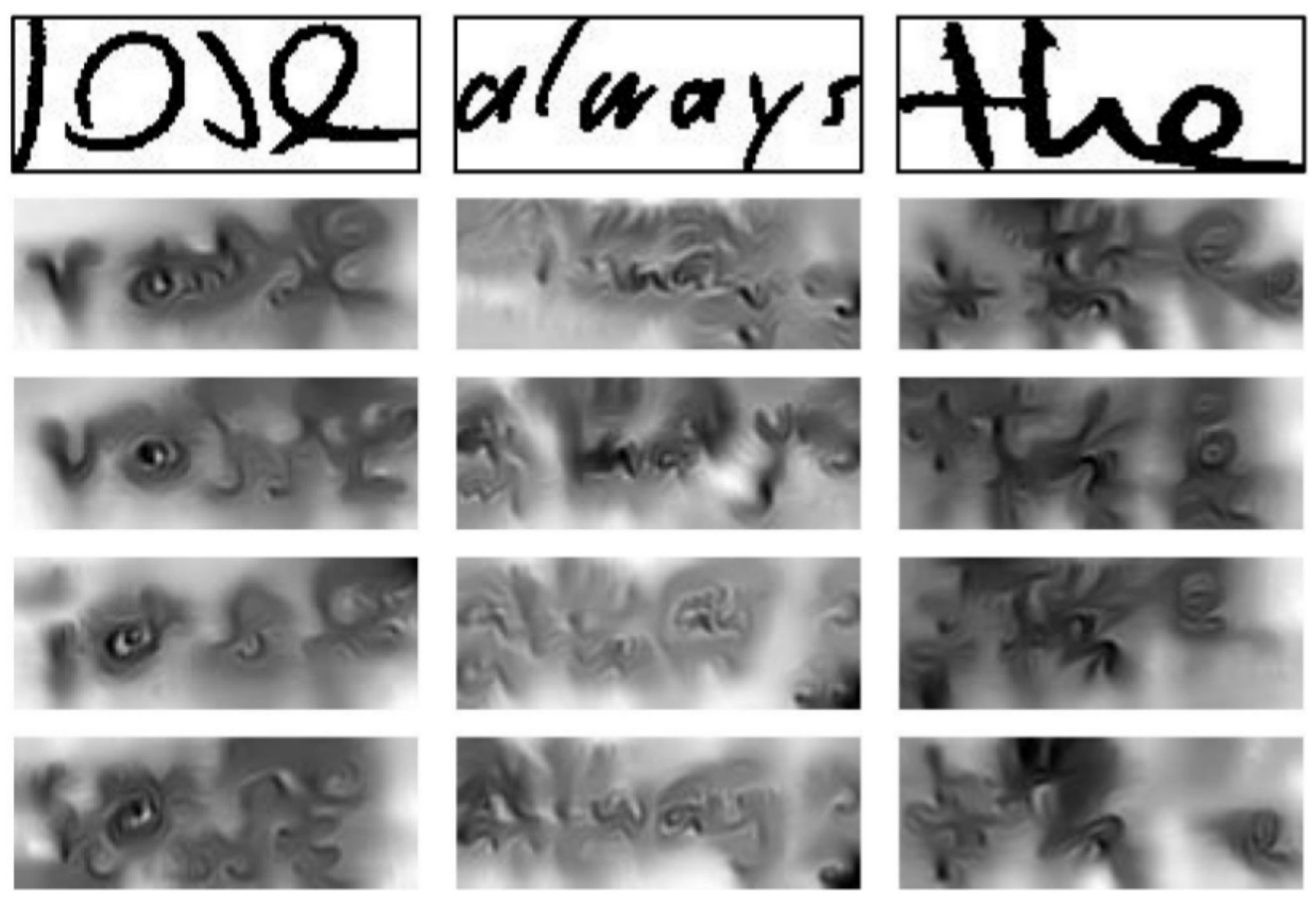}
        \caption{}
        \label{fig:lastLayer}
    \end{subfigure}
\caption{Visualizations: (a) Layer 1 weights, (b) Visualization of the strongest activation~\cite{girshick2014rich} region of a particular neuron (each column refers to one neuron) of an intermediate convolutional layer. These regions are highlighted using a bounding box inside the word image. Here we notice that, in most of the cases, each neuron focus on detecting a semantically meaningful unit, and (c) Four possible reconstructions~\cite{MahendranV15} of sample word images shown in columns. These are reconstructed from the representation obtained from the penultimate layer.}
\label{fig:visLayers}
\end{figure*}

\subsection{Curriculum Learning}
\label{subsec:currLearn}
Training with a large number of classes ($10$K for both HWNet and HWNet v2) typically results in slow convergence. To avoid such scenarios, we use the strategy from curriculum learning~\cite{Bengio2009}, where we start the training process from synthetic images showing easy examples (based on the number of characters) first and harder later. We also perform an incremental learning scheme where at initial epochs, we limit the number of classes to 500 and gradually increase the classes after achieving partial convergence. While increasing the number of classes, we copy the weights from last trained network and randomly initialize the newer weights. This improves the training process in terms of faster convergence in the presence of huge data ($\#$classes).

% \begin{figure*}[t]
% \centering
% \includegraphics[height=3cm]{Images/visLayer1.pdf}
% \includegraphics[height=3.1cm]{Images/visPatches.png}
% \includegraphics[height=3cm]{Images/lastLayer.pdf}
% \caption{Visualization: (Left) Layer 1 weight visualization, (Center) The receptive fields having the maximum activations at a particular layer and a neuron in a controlled setting of input and (Right) Four possible reconstructions~\cite{MahendranV15} of sample word images shown in columns. These are re-constructed from the representation of final layer of HWNet.}
% \label{fig:visLayers}
% \end{figure*}

\subsection{Transfer learning} 
\label{subsec:transferLearn}
It is well-known that off-the-shelf 
\textsc{cnn}s~\cite{DonahueJVHZTD14,RazavianASC14} trained for a related task could be adapted or fine-tuned to obtain reasonable and even state-of-the-art performance for new tasks. In our case, we prefer to perform transfer learning from the synthetic domain ({\sc iiit-hws}) to real world setting. In general, real world handwritten labeled corpora are not large enough to train such deep networks which contain millions of parameters and can easily over-fit on smaller datasets. Moreover, the use of synthetic corpora would give us better vocabulary coverage and also capture some frequent patterns in ngrams which commonly occur in a particular language. In our case, transfer learning achieves to reduce the domain gap between the synthetic and real world data. We employ a similar approach as presented in~\cite{Yosinski2014}, to do a careful study in transferring feature at a particular layer. The details of the study are presented in the experimental Section~\ref{subsec:transferLearnExp}.

\section{Visualizations}
\label{sec:visHWNet}
One of the most intriguing question while using a deep network is ``what is the designated behavior of a neuron trained at a particular layer?''. The question is more relevant since we are dealing with a large scale training machine with millions of parameters. Some of the recent works~\cite{szegedy2013intriguing,zeiler2014visualizing,girshick2014rich,yosinski2015understanding,MahendranV15} partly answer this question with meaningful insights on what happens behind the scenes. 

Fig.~\ref{fig:visLayers} shows the visualization of the trained HWNet v2 architecture. It is easy to visualize weights for layer 1 since the filter dimension is $64 \times 1 \times3 \times 3$ where we have: $64$ output channels (\#filters), $1$ is the channel size since the input image is gray scale, and the spatial dimension of the filter is $3\times 3$. Fig.~\ref{fig:visLayers} (a) visualizes these weights which bear a resemblance to the Gabor filters and detects edges in different orientations. %These filters are more or less similar for any deep network such as AlexNet~\cite{KrizhevskySH12}, trained on ImageNet. 
Visualizing neurons after layer 1 is non-trivial since the receptive field of these neurons keeps exploding and the filters are present in higher dimensions. Here we interpret these neurons using the non-parametric technique proposed in~\cite{girshick2014rich}, which probes from the maximum neuron activations and visualizes the receptive fields in the original image. Fig.~\ref{fig:visLayers} (b) shows such interesting patches which correspond to the maximal activation of a particular neuron taken from a convolutional layer. Here, each column corresponds to one neuron activation for different word images and sorted in a descending order as per its activation values. The first four columns of the Fig~\ref{fig:visLayers} (b) correspond to a four arbitrary neurons/channels taken from ResNet block 1, Conv. 2 (ref. Table~\ref{tab:arch} for block details) where the size of effective receptive field is $15\times 15$ pixels. The next four columns correspond to another set of four arbitrary neurons from ResNet block 2, Conv. 2 where the size of the receptive field is $37 \times 37$. The maximal activation patch is highlighted in a black box. As one can notice, each column corresponds to a semantically meaningful unit such as: Col1 neuron picks an inverted `v' sort of unit, Col3 takes wedge behavior with patches coming from letters such as `v,W,N' etc. while Col4 probes for a curve in the lower right quadrant. There also exists activation such as Row5, Col2 which does not make immediate sense in the original pixel space. The next set of columns (Col5-8) from the higher layer with more receptive field captures larger semantics such as: Col5 probes for the letter `i' and a vertical line to its right, Col6 starts detecting capital letter `A' and Col7-8 focuses partly on bigram level of information. Although we tried visualizing further layer, we couldn't make much meaningful information since, the receptive field quickly explodes to larger regions which we believe could only be explored using techniques such as~\cite{zeiler2014visualizing,yosinski2015understanding}. 

Finally, we also interpret the features which are extracted from the penultimate fully connected layer using the optimization technique proposed in~\cite{MahendranV15}. The basic idea is to invert the {\sc cnn} features back to image space and arrive at possibles images which have a high degree of probability for that encoding. This gives a better intuition of the learned layers and helps in understanding the invariances of the network. Fig.~\ref{fig:visLayers} (c) shows the possible reconstructions from three different representations. Here, we show the query images on the first row and its reconstruction in the following rows. One can observe that in almost all reconstructions, there are multiple translated copies of the characters present in the word image along with some degree of orientations. Similarly, we can see the network is invariant to the first letter being in a capital case (see Label: ``the'' at Row4, Col3) which was part of the training process. The reconstruction of the first image (see Label: ``rose'' at Row1, Col1) shows that possible reconstruction images include Label: ``rose'' (Row2, Col1) and ``jose'' (Row3, Col1) since there is an ambiguity in the query image. 

\section{Experiments}
\label{sec:exp}
In this section, we empirically evaluate the proposed word image representation, perform ablation studies to understand the importance of each architectural component in HWNet v2 and explore the usefulness of the features for the printed domain in low resource languages such as Indic scripts. To evaluate the robustness of the feature, we take the task of word spotting, where given a query image we retrieve all similar word images from a given retrieval set. 

\subsection{Datasets}
\label{subsec:datEval}
We use the four popular datasets in handwritten document analysis community out of which three are in English and one in German. Table~\ref{tab:datasets} shows different datasets and their statistics in terms of the number of words and the number of writers in case of handwritten documents. 

\renewcommand{\arraystretch}{1.4}
\setlength{\tabcolsep}{0.1em} 
\begin{table}[b]
\centering
\caption{The list of handwritten datasets used in this work. Here GW, Botany, and Konzilsprotokolle datasets are historical documents written primarily by a single author along with a few assistants(*).}
\begin{tabular}{|l|c|r|c|}\hline
\textbf{Dataset} & \textbf{Historical} & \textbf{\#Words} & \textbf{\#Writers}\\\hline
IAM & No & 1,15,320 & 657\\\hline
GW &  Yes & 4,894 & 1* \\\hline
Botany &  Yes & 20,004 & 1* \\\hline
Konzilsprotokolle (Konz.)  & Yes & 12,993 & 1* \\\hline
%IFN/ENIT &  No & 26,459 & 411 \\\hline
\end{tabular}
\label{tab:datasets}
\end{table}

\noindent \textbf{The IAM Handwriting Database~\cite{Marti02}:} It includes contributions from 657 writers making a total of 1,539 handwritten pages comprising of 115,320 words and is categorized as part of modern collection. The database is labeled at the sentence, line, and word levels. We use the official partition for writer independent text line recognition that splits the pages into training, validation, and test sets which are writer independent.

\noindent \textbf{George Washington (GW)~\cite{RathM07}:} It contains 20 pages of letters written by George Washington and his associates in 1755 and thereby categorized into historical collection. The images are annotated at word level and contain approximately 5,000 words. Since there is no official partition, we use a random set (similar to~\cite{AlmazanPAMI14}) of 75\% for training and validation and the remaining 25\% for testing.

\noindent \textbf{Botany and Konzilsprotokolle~\cite{pratikakis2016icfhr2016}:} 
These two datasets are parts of ICFHR 2016 Handwritten Keyword Spotting Competition~\cite{pratikakis2016icfhr2016}. The original competition contains data both segmentation based and free scenario. We took only the segmentation based data which contained cropped word images split into training and test sets. There were also three partitions of training sets small, medium, and large. Here we took the largest partition for conducting experiments. 

%Here we took only the largest partition which contains 16,686 training images for Botany and 9,102 for Konzilsprotokolle. And the test set contains 3,318 word images for Botany and 3,891 for Konzilsprotokolle. These two datasets are considered as part of historical document collection.

%\noindent \textbf{IFN/ENIT~\cite{pechwitz2002ifn}:} 
%The dataset contain Arabic handwritten words for city names. It consists of wide variety of writing style from nearly 411 writers and is divided into four different sets (a-d). We use the same strategy as shown in~\cite{SudholtF16} by partitioning the subsets a, b and c for training and subset d for testing. This amounts to 19,724 word images for training and 6,735 images for testing.

\subsection{Evaluation Protocol}
\label{subsubsec:wSProtocol}
For comparing results across different methods under word spotting, we use the standard information retrieval evaluation measure, mean Average Precision (mAP), which is equal to the mean area under the precision-recall curve. The selection of queries follows the protocol used in~\cite{AlmazanPAMI14}, where we filter the stopwords from the test corpus while all words (including stopwords as distractors) are kept in the retrieval dataset in which the search is performed. Our major focus is on evaluating the features in the query by example setting (\textsc{qbe}). In this setting, since the query image is taken from the corpus, the first retrieved image is not included in the mAP calculation. Since Botany and Konzilsprotokolle datasets were part of keyword spotting competition where the query and retrieval set were given independently, the dropping of query image from retrieval set was not applicable. Also, note that all evaluations for English language datasets were done in a case-insensitive manner as followed by other related works.

\subsection{Ablation Studies}
\label{subsec:ablation}
Table~\ref{tab:ablation} presents the ablation study to understand the key architectural changes and the role of data in improving the performance from HWNet baseline. These experiments are evaluated under word spotting for {\sc iam} dataset in the query by example setting. The performance of HWNet baseline architecture using just the {\sc iam} training data is reported at 0.6336. The use of residual layers gives a significant boost in performance by around $8\%$ which emphasize the generic nature of residual blocks for better learning while increasing the depth of {\sc cnn} networks. The next big improvement comes when we use variable length representation for word images under multiple scales for better coverage of scale space variation of characters written by different individuals. This is enabled by using either \textsc{roi} pooling or \textsc{tpp} before the fully connected layers. Here we observe that \textsc{tpp} performs better than \textsc{roi} since it is a generalization of \textsc{roi} in multiple scales captured in a pyramidal fashion. In our case, we essentially bring the temporal factor into account by dividing word image along the horizontal direction in each pyramid level. We now present the role of different augmentation techniques. Here we observe that using elastic and affine distortion gives around $2\%$ improvement, while the next major improvement is obtained by pre-training the network using {\sc iiit-hws} synthetic dataset. Under this setting, the network is first trained using the synthetic dataset of vocabulary 10K and later fine tuned on {\sc iam} dataset, following all architectural changes and data augmentation scheme. Finally, we also perform test time augmentation by extracting features at multiple scales. Here we resize word images at different heights (32, 48, 64) and record the maximum values of feature activation among individual scale representation. The final reported performance using HWNet v2 using \textsc{tpp} is $0.9241$.

\setlength{\tabcolsep}{5pt}
\begin{table}[t]
\centering
\caption{Ablation studies showing the effect of each of the enhancements to the baseline HWNet architecture on \textsc{iam} dataset.}
\begin{tabular}{|l|c|}\hline
\textbf{HWNet Enhancements} & \textbf{mAP} \\\hline
HWNet & 0.6336 \\\hline
HWNet+ResNet (R) & 0.7198 \\\hline
HWNet+R+Multi-Scale-ROI (ROI) & 0.8457 \\\hline
HWNet+R+Multi-Scale-TPP (TPP) & 0.8803 \\\hline
%HWNet+R+ROI+Elastic-Distortion (E) & 0.8574 \\\hline
HWNet+R+TPP+Data Augmentation (D) & 0.8996 \\\hline
%HWNet+R+M+E+IIIT-HWS (S) & 0.8973 \\\hline
HWNet+R+TPP+D+IIIT-HWS (S) & 0.9164 \\\hline
%HWNet+R+M+E+S+Test-Aug. (HWNet v2) & \textbf{0.9065}\\\hline
HWNet+R+TPP+D+S+Test-Aug. (HWNet v2) & \textbf{0.9241}\\\hline
\end{tabular}
\label{tab:ablation}
\end{table}

\subsection{Word Spotting Evaluation}
\label{subsubsec:perfEval}

\subsubsection{Architecture Evaluation}
In order to validate the efficiency of baseline HWNet~\cite{krishnan2016matching} and HWNet v2 architectures with other popular \textsc{cnn} architectures used for classification, we investigate the performance of handwritten word spotting using features obtained from two successful models, (i) AlexNet~\cite{KrizhevskySH12}, which was trained on natural images from ImageNet \textsc{lsvrc} data, and (ii) scene text recognition model (JSVZNet) trained on a large scale vocabulary of words~\cite{JaderbergSVZ14}. We validate the performance of these networks on \textsc{iam}~\cite{Marti02} dataset. Table~\ref{tab:compareModels} reports the word spotting performance for each model and compares it with HWNet. Here `Orig' refers to the model trained with its respective original datasets (e.g. AlexNet on ImageNet and JSVZNet on natural scene text) and `IAM' refers to the model fine tuned on {\sc iam } dataset. The `Orig' results (AlexNet and JSVZNet) are low, compared to the other methods for word spotting. However, they are still superior to many of the earlier handcrafted~\cite{Rodriguez-SerranoP12} features for this task. We also notice that JSVZNet performs better compared to AlexNet since it is trained for scene text words while the later model is tuned for natural scene images. The results after fine tuning (`IAM')  on \textsc{iam} dataset improves the existing results by a good margin. In the last two rows, we present the results of HWNet and HWNet v2 architectures. There is a significant improvement in results from HWNet based architectures w.r.t other architectures when trained on {\sc IAM} dataset. The last two columns of the table presents the results of only using synthetic data (IIIT-HWS) and along with fine tuning on real data (IIIT-HWS+IAM) separately. Here, the reasonable results that we obtain on just using the synthetic data (column 4) suggest the quality of generated synthetic data which captures real world variations. It also brings an interesting thought, whether in future, does such synthetic data rendering techniques limit the dependency on the availability of real data for training such systems. In section~\ref{subsec:transferLearnExp}, we present such an experiment.

\setlength{\tabcolsep}{2pt}
\begin{table}[t]
\centering
\caption{Comparative mAP evaluation of different deep networks with respect to the HWNet and HWNet v2 (TPP) network on \textsc{iam} dataset.}
\begin{tabular}{|l|c|c|c|c|}\hline
\textbf{Arch.} & \textbf{Orig} & \textbf{IAM} & \textbf{IIIT-HWS} & \textbf{IIIT-HWS+IAM} \\\hline
% AlexNet.~\cite{KrizhevskySH12} & 0.2997 & 0.4468 & * & 0.5786 \\\hline
% JSVZNet.~\cite{JaderbergSVZ14} & 0.3746 & 0.4822 & * & 0.5022 \\\hline
AlexNet & 0.2997 & 0.4468 & * & * \\\hline
JSVZNet & 0.3746 & 0.4822 & * & * \\\hline
%HWNet & * & 0.6336 & 0.5784 & 0.8219 \\\hline
HWNet & * & 0.6336 & 0.5784 & 0.8061 \\\hline
HWNet v2 & * & 0.8574 & \textbf{0.6387}
 & \textbf{0.9241} \\\hline
\end{tabular}
\label{tab:compareModels}
\end{table}

\subsubsection{State of the Art Comparison}
Table~\ref{tab:wordResults}, presents a detailed comparison between the proposed word representation and other recent methods in the task of word spotting in the query-by-example (QBE) setting on various datasets. The first three rows of the table show non-deep learning methods using engineered features. Here {\sc dtw} based method uses Vinciarelli~\cite{vinciarelli2002offline} features. The Fisher Vector ({\sc fv}) representation~\cite{PerronninR09} is computed from {\sc sift} features, reduced to 64 dimensions using {\sc pca}, and then aggregated into the Fisher Vector. Note that both {\sc dtw} based method and the Fisher representations are not learned in a supervised setting and thus cannot directly be compared to other methods which are supervised.  Here we observe that {\sc fv} performs better on these datasets compared to pixel level features using \textsc{dtw}. The attributes embedding framework described in {\sc kcsr}~\cite{AlmazanPAMI14} gives a significant boost in the performance. It also uses {\sc fv} based image representation, and projects both image and text into a {\sc phoc} word attribute space and further learns a common subspace where the correlation of both modalities is maximum. The improvement in performance emphasizes the importance of supervised learning to capture the multi-writer styles and its variations where annotated data is available.

\setlength{\tabcolsep}{1pt}
\begin{table}[t]
\centering
\caption{Quantitative evaluation of word spotting on standard handwritten datasets in query-by-example setting. Here, results for DTW and FV are taken from~\cite{AlmazanPAMI14}, while all other related works are taken from their respective papers.}
\begin{tabular}{|l|c|c|c|c|}\hline
 \textbf{Method} &\textbf{IAM} & \textbf{GW} & \textbf{Botany} & \textbf{Konz.} \\\hline
 DTW & 0.1230 &  0.6063  &  - &  - \\
 FV & 0.1566 &  0.6272 & - & -  \\
 KCSR~\cite{AlmazanPAMI14} & 0.5573  & 0.9304 & 0.7577 & 0.7791  \\
% HWNet~\cite{krishnan2016matching} & 0.8219 &  0.9484 &  0.8416 &  0.7913\\
HWNet~\cite{krishnan2016matching} & 0.8061 &  0.9484 &  0.8416 &  0.7913\\
 PHOCNet~\cite{SudholtF16} & 0.7251 & 0.9671 &0.8969 & 0.9605\\
 TPP-PHOCNet~\cite{sudholt2017evaluating}  & 0.8274 & 0.9778 & 0.9123  & 0.9770 \\
TPP-PHOCNet (BPA)~\cite{Sudholt2018}  & 0.8480 & 0.9790 & \textbf{0.9605}  & \textbf{0.9811} \\
TPP-PHOCNet (CPS)~\cite{Sudholt2018}  & 0.8274 & 0.9796 & 0.8081  & 0.9642 \\
 PHOCNet (BPA)~\cite{Sudholt2018}  & 0.8550 & 0.9758 & 0.9410  & 0.9708 \\
 Triplet-CNN~\cite{wilkinson2016semantic} & 0.8158 & 0.9800 & 0.5495  & 0.8215 \\
 %HWNet+Embed~\cite{praveenICFHR16} & 0.8425 & 0.9441 & - & - \\
 LSDE~\cite{gomezlsde2017} & - & 0.9131 & -& -\\
 \hline
 HWNet v2 (ROI) & 0.9065 & 0.9601 & 0.9401 & 0.9427\\
 HWNet v2 (TPP) & \textbf{0.9241} & \textbf{0.9824} & 0.9526 & 0.9347\\\hline
\end{tabular}
\label{tab:wordResults}
\end{table}

%\footnote{\textcolor{blue}{Note that there is an improvement in the baseline results reported for HWNet (Table~\ref{tab:wordResults} 4th row) as compared to~\cite{krishnan2016matching} since, in this work, we re-implemented the codes using PyTorch while the original HWNet~\cite{krishnan2016matching} used MatConvNet.}}

The next set of methods consider convolutional networks for extracting features optimum for word spotting. Here, we observe HWNet~\cite{krishnan2016matching} based features clearly surpasses the previous method {\sc kcsr} on all datasets. It shows the robustness of learned features using the deep network and the role of synthetic data to bootstrap the training. The next set of methods in this space use the principle of attribute embedding framework using deep {\sc cnn} networks. Here, PHOCNet~\cite{SudholtF16} and TPP-PHOCNet~\cite{sudholt2017evaluating,Sudholt2018} uses the output space of {\sc cnn} as {\sc phoc} embedding while Triplet-CNN~\cite{wilkinson2016semantic} explores with different embeddings such as {\sc phoc}, {\sc dctow} and few semantic embeddings. In the table, we report the best performance of Triplet-CNN across different proposed embeddings.  More recently, Gomez et al.~\cite{gomezlsde2017} presented a novel embedding scheme (\textsc{lsde}) by learning a subspace which respects edit distance or Levenshtein distance between a pair of samples. Although the performance is inferior from other methods, the notion of edit distance is a valid assumption while considering the string data. Finally, we compare the proposed HWNet v2 architecture on both variants (\textsc{roi, tpp}) with its predecessor (HWNet) and other methods. As we notice HWNet v2 performs significantly better for {\sc iam} and \textsc{gw} datasets where we report mAP above 0.92 and 0.98 respectively while getting comparable performance on other datasets. We would like to stress here that the boost in performance is not just because of the synthetic data, but also the architectural enhancements and the underlying formulation of learning holistic features using word classification which makes HWNet v2 different from other networks. As presented in the ablation study, in Table~\ref{tab:ablation}, one can notice the performance on {\sc iam} dataset even without adding synthetic data (IIIT-HWS), is better than other state-of-the-art networks shown in Table~\ref{tab:wordResults}.
%Note that although Triplet-CNN gives best result on {\sc gw}, it uses additional real data from CVL database to pre-train the network.

%One general observation noted using {\sc phoc} as target embedding is that, for smaller datasets and single writer scenarios such as {\sc GW} and Konzilsprotokolle these features generalize well.

Since the original HWNet v2 was trained on a large synthetic dataset, we would like to measure the performance of proposed features on out-of-vocabulary ({\sc oov}) words. Note that for the proposed method, the vocabulary comprises of a union of words present in the synthetic dataset along with the training corpus of \textsc{iam}. Here we obtain an in-vocabulary performance of 0.9223 and {\sc oov} of 0.9497 which shows the robustness of features on {\sc oov} words and also validates the unbiasedness on increasing the vocabulary size. Here, one of the justifications for the increase in {\sc oov} performance is that, in general {\sc oov} are larger words (in terms of no. of characters) which gives good contextual information to its representation and thereby easier to retrieve.

\subsubsection{Segmentation-Free Word Spotting}
\label{subsubsec:segFreeResults}
\begin{figure}[t]
\includegraphics[width=8.5cm]{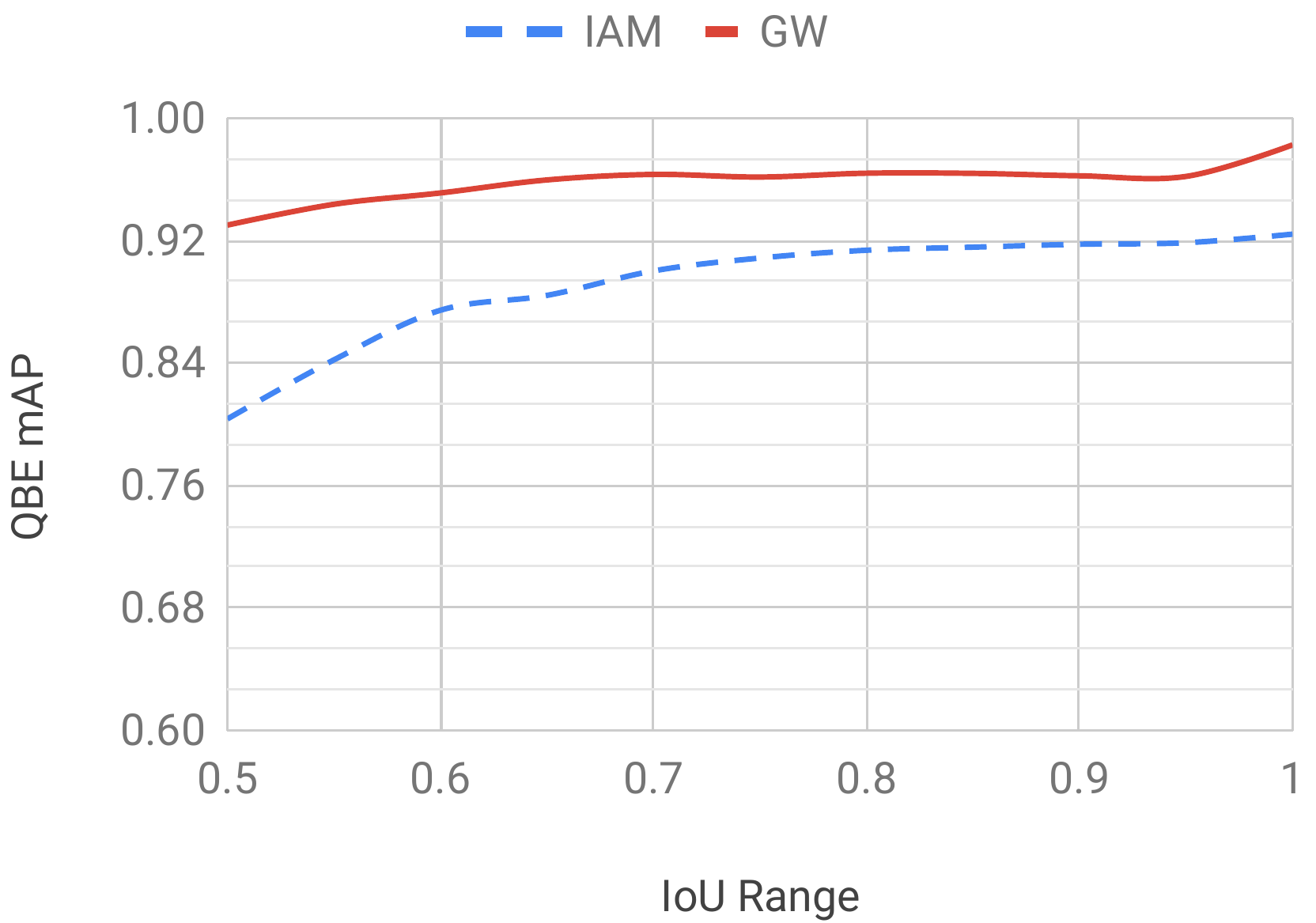}
\caption{Ablation study of word spotting in noisy word segmentation. Here we evaluate the word spotting results by perturbing the word segmentations of the test set. The perturbations are done randomly within an IoU range $(0.5,1.0)$.}
\label{fig:segFree}
\end{figure}

In this section, we evaluate the performance of HWNet v2 representation under noisy word bounding boxes and segmentation-free setting. This is in contrast to the previous evaluation where the segmentation of words are given as part of ground truth and are typically tight. In Fig.~\ref{fig:segFree}, we present an ablation study by perturbing the ground truth word segmentation of the test dataset within a certain intersection over union (IoU) range. A similar study of inaccurate cropping has also been presented in~\cite{gordo2015lewis} for word image semantic retrieval task. In our study, we vary the IoU range between $(0.5,1)$. Note that the perturbed bounding box is cropped from the original word image which could include surrounding words and noise from the page image. While the query word image is kept intact without perturbation. As shown in the figure, we observe the drop in performance marginal on increasing the IoU range for both {\sc iam} and {\sc gw} datasets. Even at an IoU range of $0.5$, we obtain a QBE mAP of $0.8034\%$ and $0.9300\%$ for {\sc iam} and {\sc gw} datasets respectively.

We now extend the evaluation to the segmentation-free scenario by computing our representation on word proposals given by an existing framework from literature. Here we use the current state of art method Ctrl-F-Net~\cite{wilkinson2017neural,wilkinsonArxiv18}. More specifically, we use the Ctrl-F-Mini model~\cite{wilkinsonArxiv18} which has a simplified architecture with better performance. The last row of Table~\ref{tab:segFreeResults}, presents the results of HWNet v2 representation as being used as the embedding instead of {\sc phoc}. As per the standard evaluation practice, we report the QBE mAP at overlap thresholds of $50\%$ and $25\%$. Here we observe a significant improvement for {\sc iam} dataset of nearly $5\%$, while for {\sc gw} dataset we obtain a comparable result. Note that, in contrast to the previous ablation study under noisy segmentation, in the segmentation-free scenario, we obtain a large number of proposals out of which many are false positives. These proposals simply act as distractors in word spotting.

\setlength{\tabcolsep}{2pt}
\begin{table}[b]
\centering
\caption{QBE mAP evaluation of HWNet v2 representation under segmentation-free scenario. Here we use the word proposals generated using the recent state of art method Ctrl-F-Mini~\cite{wilkinsonArxiv18}. As per the standard evaluation practice, we report the QBE mAP at overlap thresholds of $50\%$ and $25\%$.}
\begin{tabular}{|l|c|c|c|c|}\hline
 \multirow{2}{*}{\textbf{Embedding}} &  \multicolumn{2}{c|}{\textbf{IAM}} & \multicolumn{2}{c|}{\textbf{GW}} \\\cline{2-5}
 & 50$\%$ & 25$\%$ & 50$\%$ & 25$\%$ \\\hline
 Ctrl-F-Net (DCToW)~\cite{wilkinson2017neural} & 0.7200 & 0.7410 & 0.9050 & \textbf{0.9700} \\\hline
 Ctrl-F-Mini (PHOC)~\cite{wilkinsonArxiv18} & 0.7570 & 0.7780 & 0.9160 & \textbf{0.9700} \\\hline
 Ctrl-F-Mini (HWNet v2) & \textbf{0.8200} & \textbf{0.8240} & \textbf{0.9202} & 0.9665 \\\hline
 \end{tabular}
\label{tab:segFreeResults}
\end{table}

\subsubsection{Query-by-String Spotting Results}
\label{subsubsec:qbsResults}
In order to show the generic nature of the proposed architecture and its extension for query-by-string (QBS) spotting, in our set of parallel works~\cite{praveenICFHR16,praveenDAS18}, we have used HWNet architecture for both embedding into word attribute space defined by {\sc phoc}~\cite{praveenICFHR16} and also proposed an end2end architecture~\cite{praveenDAS18} which learns a common subspace between a text and image modality. It enables both QBE and QBS based word spotting along with word recognition using a fixed lexicon. Although the description of these works lies beyond the scope of current work, we would like to report the QBS performance reported by these architectures. Here we first report the performance of HWNet (both baseline and v2) features, embedded on to {\sc phoc} attributes space. Using the baseline HWNet architecture we obtain a QBS mAP of 0.9158~\cite{praveenICFHR16} on {\sc iam} dataset, while using the HWNet v2 architecture, we further improve this to 0.9404~\cite{praveenDAS18}. Notice that, the embedding of HWNet onto {\sc phoc} attribute space was a two-stage approach. In~\cite{praveenDAS18}, we propose an end2end scheme using HWNet v2 architecture which directly embeds both image and text into a common subspace. Here we obtain a comparable mAP of 0.9351 on {\sc iam} dataset. Please note the reported results are still competitive among the best QBS results shown in TPP-PHOCNet (CPS)~\cite{Sudholt2018} which reports an mAP of 0.9342 on {\sc iam} dataset.

\subsection{Transfer Learning}
\label{subsec:transferLearnExp}
\begin{figure}[t]
\centering
\includegraphics[width=8.5cm]{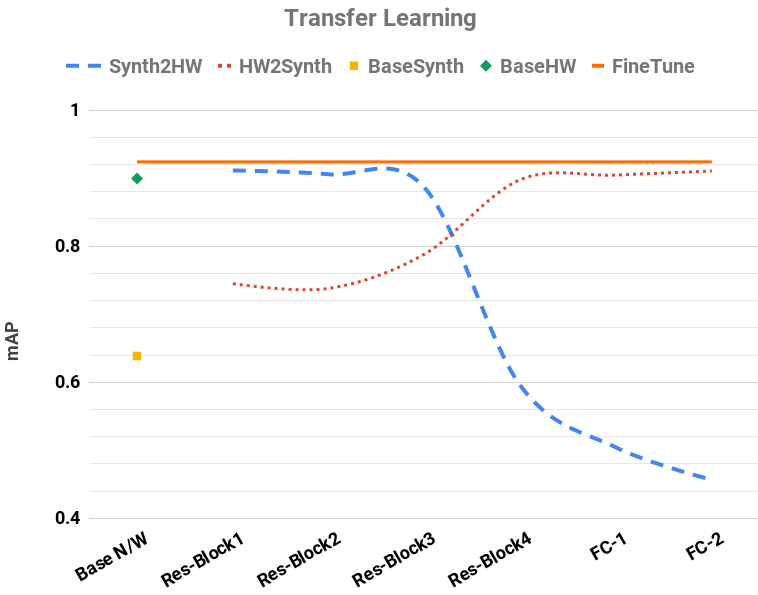}
\caption{Graph analyzing the layer for efficient transfer learning.}
\label{fig:transferLearnGraph}
\end{figure}

To better understand the transferability of features from the synthetic domain to the real domain while performing fine-tuning, we employ a similar study as presented in~\cite{Yosinski2014}. Fig.~\ref{fig:transferLearnGraph} compares the performance while doing transfer learning at different layers while keeping the layers before it either freezed or updating them. In the present analysis, we restricted the layers to four ResNet blocks and two fully connected layers immediately after it. We also analyze in three  different scenarios: (i) Synth2HW, here the base network is trained on synthetic data ({\sc iiit-hws}), and later fine-tuned on real world handwritten data {\sc iam}, (ii) HW2Synth, here the base network is trained on {\sc iam} and later fine-tuned on {\sc iiit-hws} and (iii) FineTune where all the layer weights are allowed to update while re-training. Note that, in the first two settings the weights before the particular ResNet block/FC layer is kept freezed. We also report the base network performance while training on Synth and \textsc{iam} datasets. The performance of Base Synth network is quite less, which indicates the presence of domain gap while Base HW performs fairly good with the enhancements done as part of HWNet v2. Under the three settings of transfer learning, we notice that the best performance is obtained in the third setting of fine tuning where all the weights are simultaneously updated without freezing any layers. Except for the setting of Synth2HW where we keep block 1 freezed, in all other scenarios we obtain inferior results which suggest that initial layer filters learned by the synthetic data are reasonably robust and generalize well to real scenarios.

\setlength{\tabcolsep}{1pt}
\begin{table}[b]
\centering
\caption{Evaluation of word spotting using mAP on {\sc iam} test dataset by training HWNet v2 with entire synthetic dataset while fine tuning on varying percentage of {\sc iam} training data. Here Train=0.0\% refers only using synthetic data for training.}

\begin{tabular}{|c|c|c|c|c|c|c|c|}\hline
Train $\%$ & 1.0 & 0.8 & 0.6 & 0.4 & 0.2 & 0.1 & 0.0\\\hline
%HWNet & 0.8219 & 0.8313 & 0.8235 & 0.8090& 0.7677 & 0.7333 & 0.5760\\\hline
HWNet v2 & 0.9241 & 0.9150 & 0.9084 & 0.9025 & 0.8773 & 0.8625 & 0.6387\\\hline
\end{tabular}
\label{tab:trainPercent}
\end{table}

In Table~\ref{tab:trainPercent} we present an interesting outcome of transfer learning from synthetic data. Here we experiment the reduced need for real data for training HWNet v2 architecture by varying the amount of training data as compared to previous experiments. Here also we take {\sc iam} dataset as our test bench and use a different proportion of real data and compare the performance with architecture which uses full training data. Here full is depicted as 1.0 while 0.0 depicts the use of only synthetic data. Note that all these experiments are first trained on entire synthetic data and later fine-tuned using varying proportions. As one can notice the drop in performance with reduced real training data starts very slowly and surprisingly even using a mere 10\% of real data only drops the performance by 6\%. Although this suggests the lesser dependency on real data, we believe this needs a thorough study (out of the scope of current work) to evaluate the differences in domain gap between synthetic data and the target handwritten styles.

\subsection{t-SNE Embedding}
\label{subsec:tSNE}
\begin{figure}[b]
\includegraphics[width=8.5cm]{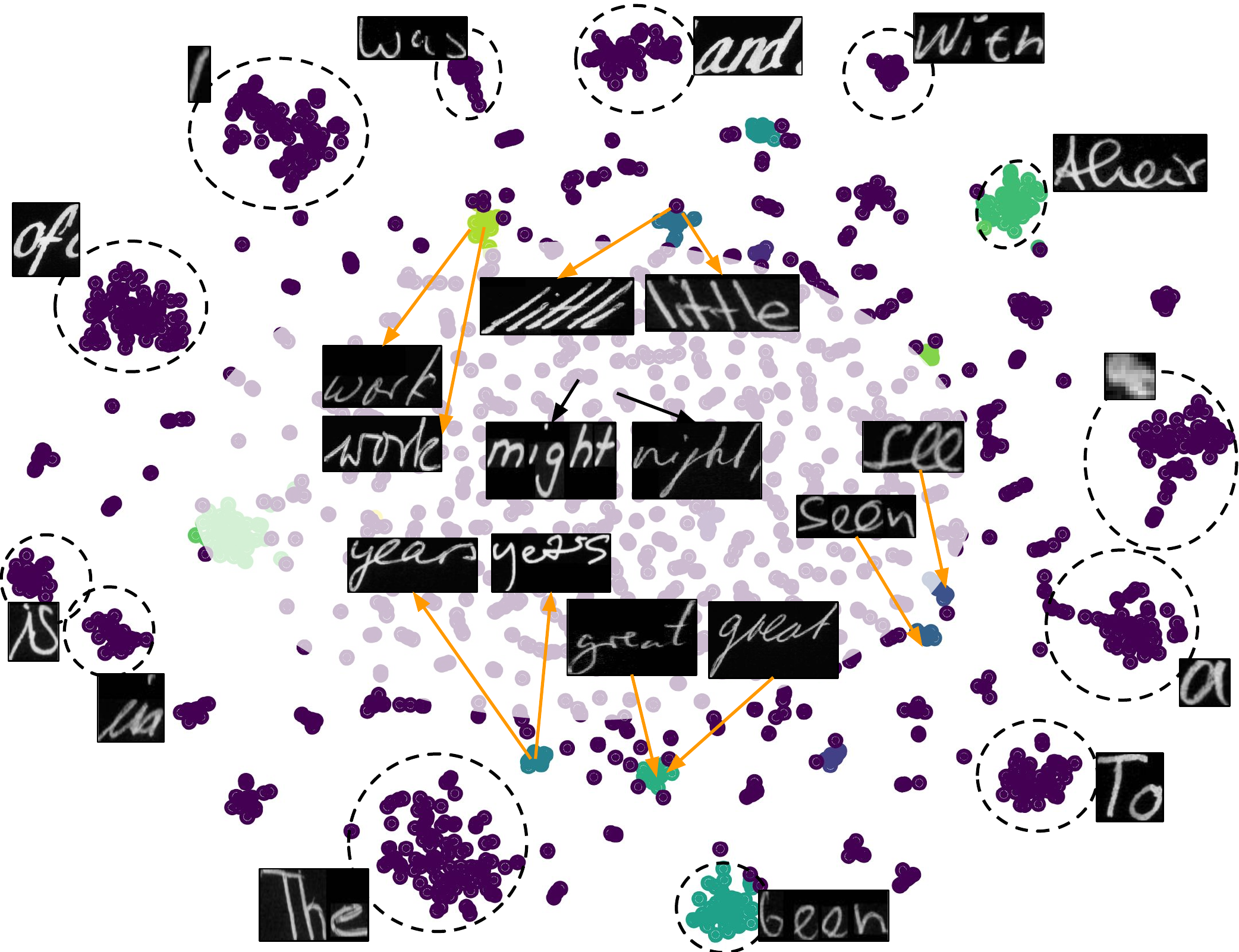}
\caption{t-SNE embedding of word image representation taken from the validation set from \textsc{iam} dataset.}
\label{fig:tsne}
\end{figure}

To better understand the learned representation space, in Fig.~\ref{fig:tsne} we present the t-SNE~\cite{maaten2008visualizing} embedding of word image representation for visualization of higher dimensional feature space. We took the validation set from \textsc{iam} dataset for t-SNE embedding. The visualization shown here brings interesting insights where we see that the stopwords which occur higher in number are shifted to the periphery of the space where they neatly group into tight clusters. The rest of the significant words from the vocabulary are scattered inside the region. On a closer look, we observe the neighboring sample points belong to the same word or words which are lexically near in space. Please note in the inner region, it is difficult to make out tight clusters since the frequency of these words is quite low. Few of such points are presented along with the actual validation word image taken from the dataset. One can notice the invariance in representation space which clearly take lexical content into account. We also observe smoothness in the representation space where words such as (might, night) and (see, seen) are found closer in the embedding.

\begin{figure}[t]
\includegraphics[width=8.5cm]{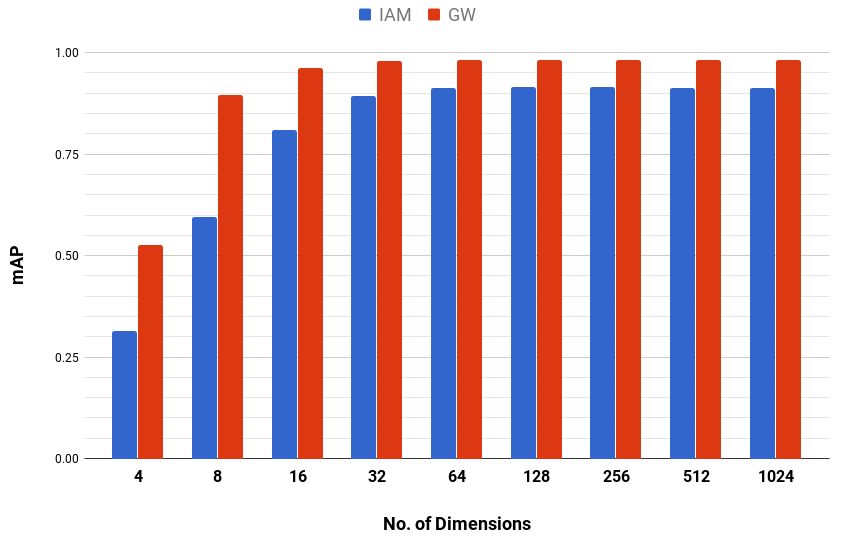}
\caption{Evaluation of compression of learned representation using \textsc{pca} on \textsc{iam} and \textsc{gw} datasets.}
\label{fig:pca}
\end{figure}

\begin{figure*}[t]
\centering
\includegraphics[width=17.5cm]{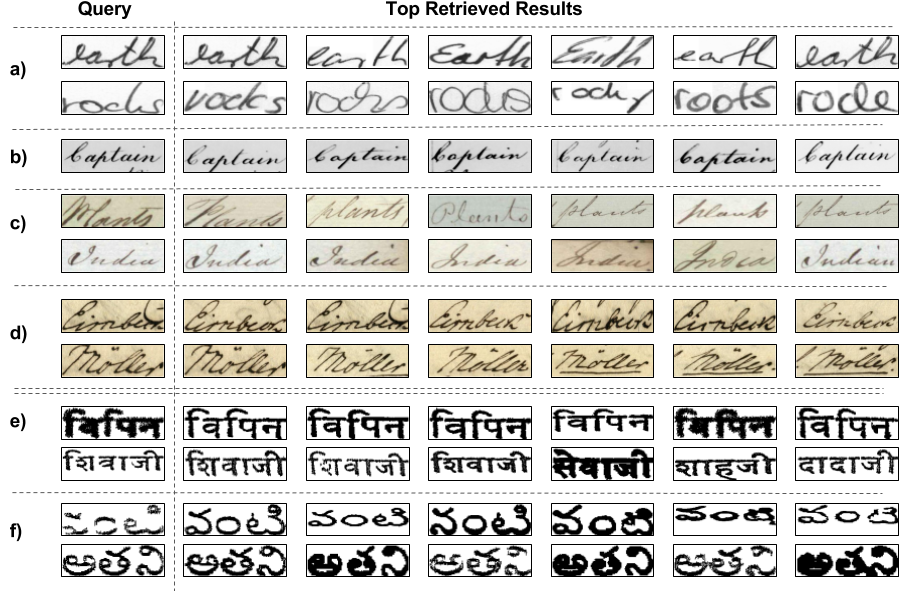}
\caption{Qualitative results of word spotting. In each row, the leftmost image is the query and remaining are the top retrieved word images based on features similarity. Here (a-d) refers to images taken from \textsc{iam}, \textsc{gw}, Botany and Konzilsprotokolle  datasets respectively. The last two rows (e-f) shows results from printed dataset containing degraded Hindi and Telugu books of the \textsc{dli} corpus respectively.}
\label{fig:wordSpot}
\end{figure*}

\subsection{Compression of Representation}
\label{subsubsec:compress}
%need for efficient codes; connect with tsne; linear dimensionality reduction with PCA; analysis of the graph
Taking inspirations from our previous analysis on embedding the representation onto a two dimensional space where we gained interesting insights on the similarity of neighbors even in extreme form of compression. We now formally extract a lower dimensional representation of HWNet features using principal component analysis (\textsc{pca}), which is a popular linear dimensionality reduction technique. We use the validation data to extract the top eigen vectors in the representation space and project each of the test data using them. Fig.~\ref{fig:pca} shows the performance variation across different compression levels starting from 4 to 1024 on both \textsc{iam} and \textsc{gw} datasets. Note that original HWNet features are of size 2048. It is interesting to see that there is minimal or no drop in performance in the range of dimensions 32-1024 and we even get a minor improvement in performance for 128 dimensions. This clearly states that the original HWNet network is able to capture non-linear relationships in the data and the final representation space only contains linear components. We obtain an mAP of 0.8942 on \textsc{iam} using 32 dimensions which is still the state-of-the-art as compared to other related methods reported in Table~\ref{tab:wordResults}. Note that the performance obtained using mere 8 dimensions (0.5942) is still better than some of the non-deep learning methods. 

%The analysis of compression is particularly important from the perspective of large scale storage and retrieval of word images which is quite common in digital libraries.

\begin{figure}[t]
\includegraphics[width=8.5cm]{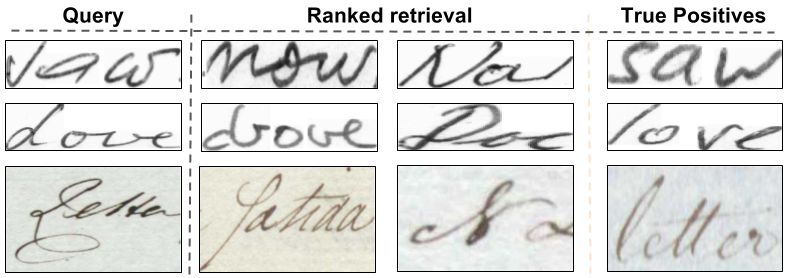}
\caption{Sample failure case images where the representation fails to match lexically correct neighbors.}
\label{fig:fail-case}
\end{figure}

\subsection{Qualitative Results and Failure Scenarios}
\label{subsubsec:qualFail}
Fig.~\ref{fig:wordSpot} shows sample qualitative results obtained under different datasets in the query by example setting using HWNet v2 features. One can notice the robustness of features which make it invariant across different writer variations in {\sc iam}, word capitalization forms, and common degradations, as seen in the historical datasets such as Botany and Konzilsprotokolle etc. Fig.~\ref{fig:fail-case} presents some of the failure cases where our representation fails to match lexically correct neighbors. Here we show the query image and the top nearest neighbors in the representation space which is incorrect. We also show a sample true positive word image which was lying farther in the representation space. Here top two rows show an inherent ambiguity which was created due to the complexity of handwriting where word ``saw'' and ``love'' are being retrieved as ``now'' and ``dove''. In the case of the third row, we observe that stylization of the character `L' created a large impact. In summary, we find the ambiguity/complexity of writing and fine grained similarity in different words to be the major causes of failures.

%while the issue in the last row of examples which is written in Arabic language is fine grained change which is even hard to detect for a non-native writer

\subsection{Applicability for Printed Documents}
\label{sec:printDocs}
In order to validate the performance of the proposed architecture on a different modality such as printed documents, we tested our architecture on a standard book from English and few challenging books taken from DLI corpus~\cite{DLI} in Hindi and Telugu. 

\noindent \textbf{English-1601~\cite{Yalniz12}:} This is a book in English titled ``Adventures of Sherlock Holmes'' written by Arthur Conan Doyle. This was first used in~\cite{Yalniz12} for comparing \textsc{ocr} based results with image search. 
%We use this corpus with train-val-test splits of 60-20-20$\%$ ratio.

\setlength{\tabcolsep}{5pt}
\begin{table}[t]
\centering
\caption{The list of printed datasets used in this work. Here both Hindi and Telugu datasets are taken from Digital Library of India~\cite{DLI} corpus.}
\begin{tabular}{|l|r|r|}\hline
\textbf{Dataset} &  \textbf{\#Pages} & \textbf{\#Words} \\\hline
English-1601 & 310 & 1,13,008  \\\hline
DLI Hindi (HS1) & 1,533 & 4,20,100 \\\hline
DLI Telugu (TS1) & 1,005 & 1,61,265 \\\hline
\end{tabular}
\label{tab:printedDatasets}
\end{table}
\setlength{\tabcolsep}{5pt}

\noindent \textbf{DLI Hindi and Telugu~\cite{krishnanICVGIP12}:} 
These two datasets, belonging to Hindi and Telugu languages from Indic scripts are part of Digital Library of India (DLI)~\cite{DLI} project. DLI has emerged as one of the largest collections of document images in Indian scripts. Many of the pages present in DLI contains serious forms of document degradation which restricts present day \textsc{ocr}'s and text spotting systems to work efficiently. We take one such subset~\cite{krishnanICVGIP12} which was annotated at the level of lines and words and referred to as HS1 and TS1 datasets. 
%Similar to English-1601, we split the annotation into train-val-test with 60-20-20$\%$ ratio.

\setlength{\tabcolsep}{0.2em} 
\begin{table}[b]
\centering
\caption{Quantitative evaluation of word spotting on printed datasets.}
\begin{tabular}{|l|c|c|c|c|}\hline
\textbf{Method} & \textbf{Supervision} & \textbf{English} & \textbf{Hindi} & \textbf{Telugu} \\\hline
Yalniz et.al~\cite{Yalniz12} & No & 0.9300 & - & - \\\hline
Krishnan et.al~\cite{krishnanICVGIP12} &No& - & 0.6055 & 0.7438 \\\hline
HWNet v2 (TPP) &Yes& \textbf{0.9570} & \textbf{0.9509} & \textbf{0.9582} \\\hline
\end{tabular}
\label{tab:printedResults}
\end{table}

We present the results in Table~\ref{tab:printedResults} where we compare HWNet v2 (TPP) with previous methods. Note that the previous methods, Yalniz et al.~\cite{Yalniz12} and Krishnan et al.~\cite{krishnanICVGIP12} use features from \textsc{bow} pipeline which is essentially unsupervised in nature and not directly comparable with neural codes which are supervised. We would like to contrast the advantages of supervised learning and also baseline  the performance on these datasets with deep features. Here for English dataset, we just use the pre-trained model from \textsc{iiit-hws} without any fine-tuning, while for Hindi and Telugu datasets, we perform fine-tuning on the training corpus of these datasets respectively. From the results, we notice that the performance of word spotting has improved significantly on all these datasets and for printed English, one could directly use HWNet v2 as off-the-shelf for various document tasks. The improvement of results in Hindi and Telugu also suggests that such an architecture can be used for various languages with wide variations in scripts and language constructs. One can also notice the amount of degradation in the retrieved words from the qualitative results shown in the last two rows of Fig.~\ref{fig:wordSpot} which indicates the robustness of the proposed features.

\subsection{Implementation Details}
\label{subsec:implDetails}
HWNet v2 network is trained using stochastic gradient descent algorithm with momentum. We set the momentum factor to be $0.9$ and the learning rate is set as $1e-2$ while training from scratch on synthetic data. In the case of fine tuning, the learning rate is initialized from $1e-3$ and manually reduced by a factor of 10 once the loss does not change within a certain threshold in last five epochs. 
The weights are initialized using He initialization\cite{he2015delving}. With respect to data augmentation, we perform on the fly augmentation with 50\% probability whether to augment the current sample from the mini-batch. For elastic distortion, we set the hyper-parameters $\alpha=0.8, \sigma=0.08$ denoted as scaling and smoothing parameters~\cite{SimardSP03}, which regulates the amount of distortion to apply. For affine transformation, we randomly pick whether to rotate, shear or pad. The rotation and shear angles are sampled in the range of $(-5,15)$ and $(-0.5,0.5)$ degrees respectively. For bringing translation in-variance, we randomly  insert padding in the four boundaries within a range of 0-20 pixels. In our experiments for the segmentation-free scenario, while training the models we also perturbed the ground truth bounding boxes of training word images within an IoU range of $(0.75,1.0)$ to learn features robust against the noise while testing.

% \subsubsection{Computational Analysis}
% \label{subsubsec:compAnal}
We use NVIDIA GeForce GTX 1080 Ti GPUs for all our experimentation and the codes are written using PyTorch 0.2 library. On a batch size of 8, our code takes GPU RAM of around 6.5GB and 1.4GB for training and inference respectively. The model roughly takes around 10 milliseconds for computing representation for each word image.

\subsection{Release of Datasets and Codes}
\label{subsec:codes}
For the reproducibility of research and use of HWNet v2 features for different document processing tasks, we have released the {\sc iiit-hws} synthetic dataset, pre-trained network files of HWNet v2 architecture along with codes for extracting features for word images. More details are shared in the project page\footnote{\url{http://cvit.iiit.ac.in/research/projects/cvit-projects/hwnet}}.
%We also provide support for OpenCV platform using C++ backend for developing products and solutions.

\section{Conclusion and Future Work}
\label{sec:conc}
We introduce a generic deep convolutional framework for learning word image representation for document images. The underlying architecture HWNet v2 uses synthetic data for efficient pre-training and also uses various data augmentation schemes which mimic the natural process of text creation in documents. We further provide different insights into the fine tuning process and also understand the invariances learned at various layers using some recent visualization techniques. We successfully demonstrate the robustness of learned representation in terms of both performance and dimensionality of final representation, on challenging historical manuscripts in both handwritten and printed modalities. 
%The features computed from baseline HWNet architecture have already shown to work well for word recognition tasks as presented in~\cite{praveenICFHR16}.

As part of future work, we would like to explore the granularity of representation at the level of patches by augmenting the final representation with intermediate layer activations. We believe such an enriched representation would capture local information which would be useful to distinguish between classes with minimum edit distance. We are also exploring the use of attribute~\cite{AlmazanPAMI14} information as conditional data while learning the final representation. In this direction, our initial results which has been accepted in~\cite{praveenDAS18} looks promising for word spotting.

\bibliographystyle{spmpsci}      % mathematics and physical sciences
\bibliography{praveenIJDAR}   % name your BibTeX data base

\end{document}